\documentclass{article}

\PassOptionsToPackage{numbers}{natbib}
\usepackage[preprint]{neurips_2026}

\usepackage[utf8]{inputenc}
\usepackage[T1]{fontenc}
\usepackage{hyperref}
\usepackage{url}
\usepackage{xcolor}
\usepackage{colortbl}
\usepackage{multirow}
\usepackage{graphicx}
\usepackage{booktabs}
\usepackage{amsmath}
\usepackage{amsfonts}
\usepackage{nicefrac}
\usepackage{microtype}
\usepackage{marvosym}
\usepackage{float}
\usepackage{placeins}

\newcommand{\corremail}{\textsuperscript{\raisebox{-0.15ex}{\scriptsize\Letter}}}
\definecolor{citeblue}{RGB}{40,90,160}

\hypersetup{
    colorlinks=true,
    citecolor=citeblue,
    linkcolor=black,
    urlcolor=black
}

\title{PriorVLA: Prior-Preserving Adaptation for Vision-Language-Action Models}

\author{
  \textbf{Xinyu Guo}$^{1,3,4,\dagger}$ \quad
  \textbf{Bin Xie}$^{2,\ddagger}$ \quad
  \textbf{Wei Chai}$^{5,\dagger}$ \quad
  \textbf{Xianchi Deng}$^{2}$ \\
  \textbf{Tiancai Wang}$^{2}$ \quad
  \textbf{Zhengxing Wu}$^{1}$\corremail \quad
  \textbf{Xingyu Chen}$^{4}$\corremail \\[0.45em]
  {\normalfont\small
  \begin{tabular}{c}
  $^1$Institute of Automation, Chinese Academy of Sciences \quad
  $^2$Dexmal \\
  $^3$University of Chinese Academy of Sciences \quad
  $^4$Zhongguancun Academy \\
  $^5$Nanjing University of Aeronautics and Astronautics
  \end{tabular}
  }
}

\begin{document}
\maketitle
\vspace{-1.5em}

\begingroup
\renewcommand{\thefootnote}{}
\footnotetext{
$^\dagger$ Work done during an internship at Dexmal.
\quad
$^\ddagger$ Project leader.
\quad
\corremail~Corresponding author.
}
\addtocounter{footnote}{-1}
\endgroup

\begin{abstract}
Large-scale pretraining has made Vision-Language-Action (VLA) models promising foundations for generalist robot manipulation, yet adapting them to downstream tasks remains necessary.
However, the common practice of full fine-tuning treats pretraining as initialization and can shift broad priors toward narrow training-distribution patterns.
We propose PriorVLA, a novel framework that preserves pretrained priors and learns to leverage them for effective adaptation.
PriorVLA keeps a frozen Prior Expert as a read-only prior source and trains an Adaptation Expert for downstream specialization.
Expert Queries capture scene priors from the pretrained VLM and motor priors from the Prior Expert, integrating both into the Adaptation Expert to guide adaptation.
Together, PriorVLA updates only 25\% of the parameters updated by full fine-tuning.
Across RoboTwin 2.0, LIBERO, and real-world tasks, PriorVLA achieves stronger overall performance than full fine-tuning and state-of-the-art VLA baselines, with the largest gains under out-of-distribution (OOD) and few-shot settings.
PriorVLA improves over $\pi_{0.5}$ by 11 points on RoboTwin 2.0-Hard and achieves 99.1\% average success on LIBERO.
Across eight real-world tasks and two embodiments, PriorVLA reaches 81\% in-distribution (ID) and 57\% OOD success with standard data.
With only 10 demonstrations per task, PriorVLA reaches 48\% ID and 32\% OOD success, surpassing $\pi_{0.5}$ by 24 and 22 points, respectively.
Project Page: \href{https://priorvla.github.io/}{\textcolor{citeblue}{https://priorvla.github.io/}}
\end{abstract}

\section{Introduction}
Vision-Language-Action (VLA) models have emerged as promising foundations for generalist robot manipulation~\cite{rt1,rt2,openvla,pi0,pi05}.
Through large-scale pretraining on diverse robot data, a single policy can learn broad priors that support general manipulation capabilities, grounding language instructions in visual observations and generating actions across tasks and embodiments~\cite{bridge_data_v2,open_x_embodiment,droid,octo,gr2,univla_latent}.
Yet the complexity and diversity of real-world manipulation cannot be fully covered by pretraining alone, making downstream adaptation necessary~\cite{openvla_oft,cogact,spatialvla,tinyvla,ki,vla_adapter}.
In practice, downstream adaptation is commonly performed via full fine-tuning, which can quickly improve in-distribution (ID) performance~\cite{openvla_oft,pi05}.
However, full fine-tuning treats the pretrained model mainly as an initialization and updates all parameters to fit limited task-specific data, which can shift broad pretrained priors toward narrow training-distribution patterns and weaken out-of-distribution (OOD) generalization~\cite{pi05,openvla_oft,maps,retain,ki}.
Under this paradigm, generalizable adaptation requires additional diverse demonstrations to cover broader variations~\cite{mimicgen,robocasa,robotwin2,roboverse}, which is a poor use of large-scale pretraining.

\begin{figure}[!t]
    \centering
    \includegraphics[width=1\linewidth, trim={115pt 62pt 130pt 95pt}, clip]{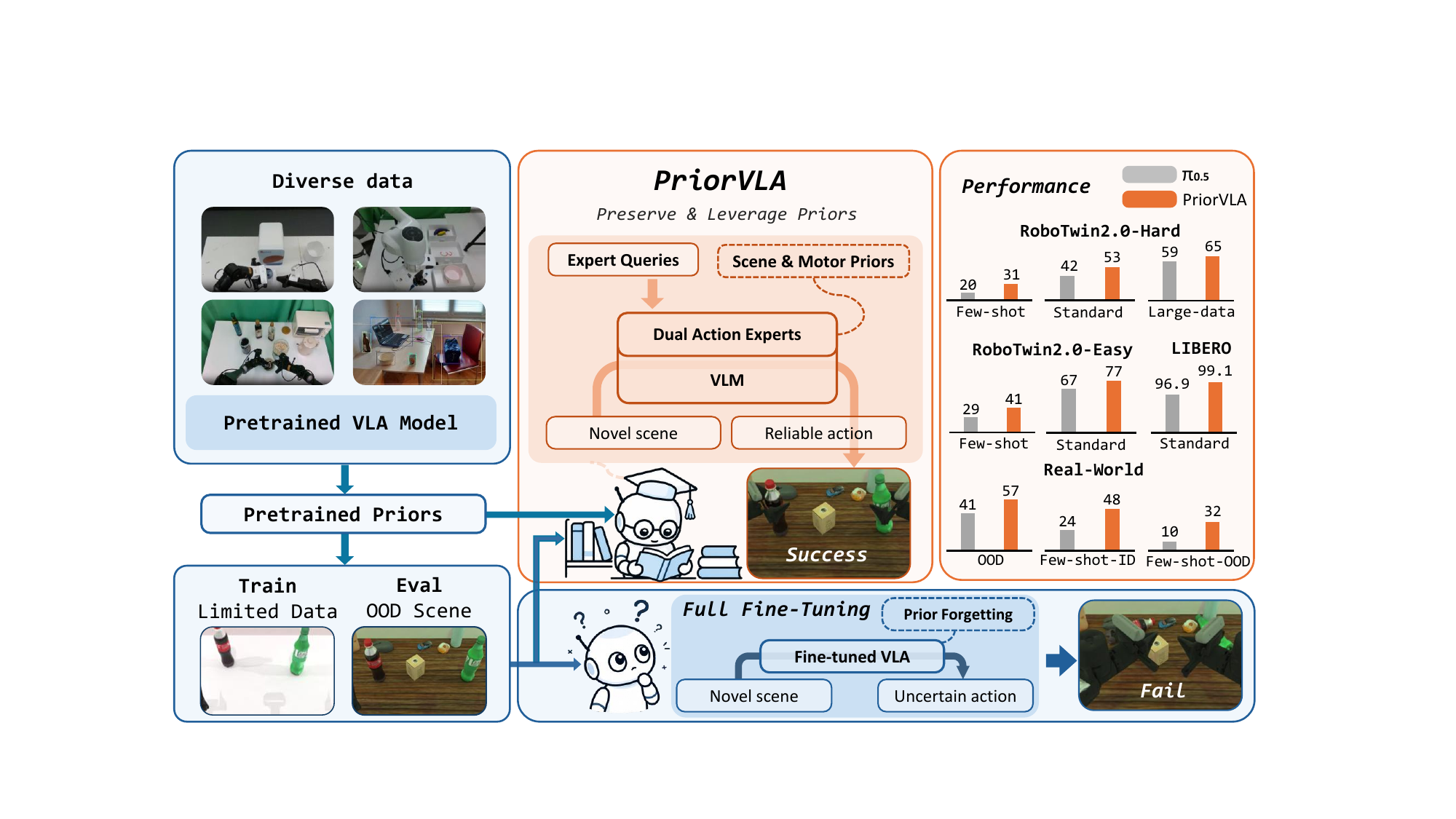}
\caption{
\textbf{Overview of PriorVLA.}
Large-scale pretraining provides broad priors for general manipulation, but full fine-tuning on limited downstream data can treat these priors mainly as initialization and lead to prior forgetting, especially when evaluated under OOD scenes.
PriorVLA instead preserves and leverages pretrained scene and motor priors through Dual Action Experts and Expert Queries, improving adaptation across simulation and real-world tasks with the largest gains in few-shot and OOD settings.
}
    \label{fig:overview}
\vspace{-5.5mm}
\end{figure}

Large-scale VLA pretraining is valuable not only as an effective initialization for downstream fine-tuning, but also as a source of broad priors for generating robot actions from visual observations and language instructions across tasks and embodiments.
Effective VLA adaptation should therefore preserve these priors and make them usable for downstream tasks.
These priors are not explicit rules or final action outputs, but emerge in the pretrained model's forward pass.
They provide complementary scene and motor priors: the VLM encodes task-relevant visual structure, while the action expert encodes action-generation regularities.
These priors supplement limited downstream data with pretrained structure, helping the adapted policy handle novel observations and generate reliable actions in few-shot and OOD settings.
Existing adaptation methods either reduce or constrain parameter updates~\cite{lora,openvla_oft,maps,retain}, or bridge frozen vision-language features to action policies~\cite{ki,vla_adapter}.
However, they primarily focus on updating, restricting, or connecting pretrained components, rather than explicitly preserving pretrained forward-pass representations as usable prior sources.
What remains missing is a mechanism that preserves these priors and provides learnable interfaces for the adapted policy to use them.

To this end, we propose \textbf{PriorVLA}, a prior-preserving adaptation framework that preserves pretrained priors and learns to leverage them during downstream adaptation, as summarized in Fig.~\ref{fig:overview}.
PriorVLA introduces Dual Action Experts to separate prior preservation from task specialization.
The original pretrained action expert is kept as a frozen Prior Expert, while a trainable Adaptation Expert, initialized from the same weights, specializes to the downstream task.
The Prior Expert is not used for its final action prediction; instead, it serves as a read-only forward path whose internal representations provide motor priors from pretraining.
PriorVLA further introduces Expert Queries as learnable interfaces for using preserved priors.
Scene Queries capture task-relevant scene priors from the pretrained VLM, Motor Queries capture motor priors from the Prior Expert, and Action Queries integrate these priors inside the Adaptation Expert to guide action generation.
Although trained with the same action objective as full fine-tuning, PriorVLA changes what adaptation learns: it learns to capture and integrate preserved priors, rather than adapting by shifting them toward narrow training-distribution patterns.

We conduct extensive evaluations of PriorVLA across simulation and real-world settings, including RoboTwin 2.0~\cite{robotwin2}, LIBERO~\cite{libero}, and eight real-world tasks spanning two embodiments: a single-arm Franka robot and a dual-arm AC-One robot platform.
PriorVLA achieves stronger overall performance than full fine-tuning and state-of-the-art VLA baselines, with the largest gains under out-of-distribution evaluation and limited downstream supervision.
On RoboTwin 2.0-Hard, PriorVLA surpasses $\pi_{0.5}$ by 11 points, demonstrating stronger OOD generalization.
On LIBERO, it achieves 99.1\% average success across four task suites.
Across eight real-world tasks, PriorVLA reaches 81\% ID and 57\% OOD success in the standard-data setting.
With only 10 demonstrations per task, PriorVLA reaches 48\% ID and 32\% OOD success, surpassing $\pi_{0.5}$ by 24 and 22 points, respectively.
These gains are achieved while updating only 25\% as many parameters as full fine-tuning.

Our contributions are summarized as follows:
\begin{itemize}
    \item We motivate a prior-preserving view of downstream VLA adaptation: beyond serving as initialization, large-scale pretraining provides forward-pass priors that should be preserved and leveraged during adaptation. This reframes adaptation as learning to use pretrained structure, rather than only fitting limited downstream data.
    
    \item We propose PriorVLA, a prior-preserving adaptation framework with two coupled designs: Dual Action Experts separate prior preservation from task specialization, and Expert Queries capture and integrate preserved scene and motor priors for action generation.

    \item We demonstrate stronger overall performance across RoboTwin 2.0, LIBERO, and eight real-world tasks on two embodiments, with the largest gains in OOD and few-shot settings, while updating only 25\% as many parameters as full fine-tuning. Ablations further show that the preserved pretrained priors, rather than simply adding frozen branches or reducing trainable parameters, are critical to these gains.
\end{itemize}

\section{Related Work}
\label{sec:related}
\paragraph{Pretrained VLA Models.}
Robotic manipulation policies trained on narrow task-specific datasets often struggle to generalize beyond their training distributions~\cite{diffusion_policy, act,zheng2025densegrounding, peract, rvt,shi2026spatialactor}.
Recent Vision-Language-Action (VLA) models address this limitation by scaling robot data, simulation experience, and pretrained vision-language backbones.
Large-scale real-robot datasets and synthetic data generators have substantially expanded manipulation coverage across tasks, scenes, objects, and embodiments~\cite{rt1, bridge_data_v2, open_x_embodiment, droid, agibot_world, mimicgen, robocasa, robotwin2, roboverse, autotrialgen}.
Built on these data sources, RT-2, OpenVLA, and $\pi_0$ demonstrate increasingly generalist robot policies through large-scale pretraining~\cite{rt2, openvla, pi0}, while recent systems further improve action modeling, data mixtures, and embodied reasoning~\cite{pi05, rdt, groot_n1, gemini_robotics15, smolvla}.
These advances make pretrained VLAs valuable foundations for downstream manipulation, as many pretrained capabilities are difficult to recover from limited task-specific data.

\paragraph{Downstream Adaptation of VLA Models.}
Adapting pretrained VLAs to new tasks, robots, and data regimes is still challenging: full fine-tuning can quickly improve in-distribution performance, but may over-specialize the model and weaken pretrained capabilities.
Recent work improves VLA adaptation by refining action representations and fine-tuning objectives~\cite{openvla_oft}, reducing trainable parameters with low-rank updates~\cite{lora}, or freezing/protecting the VLM while placing more adaptation capacity in action-side modules~\cite{groot_n1, ki}.
Other approaches improve robustness by constraining or merging parameter updates~\cite{maps, retain}, incorporating temporal context~\cite{hamlet}, or preserving previous capabilities through continual-learning regularization and replay~\cite{ewc, gem}.
Lightweight interface designs further connect vision-language representations to action policies through query-style tokens or bridge attention~\cite{openvla_oft, vla_adapter}.
RICL~\cite{ricl} and ICRT~\cite{icrt} use demonstration context for task adaptation, whereas PriorVLA learns an adapted policy without demonstration prompts at inference.
How to preserve and leverage pretrained priors during downstream action generation remains less explored in current VLA adaptation frameworks.

\section{Method}
\label{sec:method}
\subsection{Problem Formulation}
\label{sec:problem_formulation}

We consider a vision-language-action (VLA) policy that predicts a future action chunk for robot manipulation. Given visual observation $o_t$, language instruction $l$, and proprioceptive state $s_t$, the policy outputs an action chunk $\mathbf{a}_{t:t+H-1}$, where $H$ denotes the action horizon.

A common pretrained VLA instantiation couples a vision-language model (VLM) with a flow-matching-based action expert (AE)~\cite{pi0, pi05}. During denoising, the policy conditions on a noisy action chunk $\tilde{\mathbf{a}}_{t:t+H-1}$:
\begin{equation}
    \hat{\mathbf{a}}_{t:t+H-1}
    =
    \pi_{\theta}\!\left(
    \tilde{\mathbf{a}}_{t:t+H-1},\,o_t,\,l,\,s_t
    \right),
\end{equation}
where $\hat{\mathbf{a}}_{t:t+H-1}$ denotes the denoising prediction. Through large-scale pretraining, the VLM acquires general vision-language representations, while the AE learns reusable motor priors for action generation. These complementary pretrained capabilities form the basis for downstream adaptation.

\begin{figure}[!t]
    \centering
    \includegraphics[width=1\linewidth, trim={45pt 115pt 82pt 105pt}, clip]{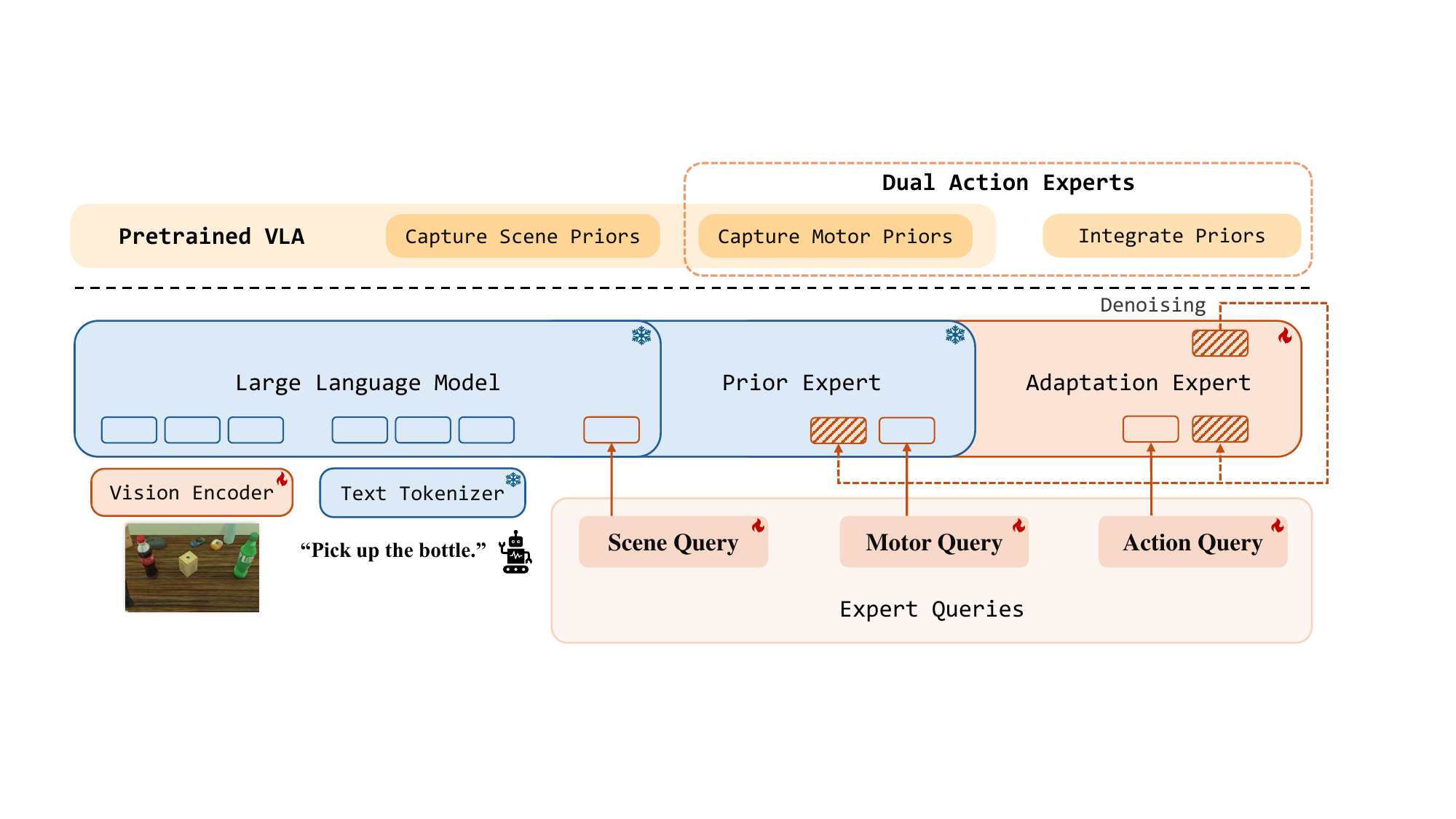}
    \caption{
    \textbf{PriorVLA architecture.}
    PriorVLA builds on a pretrained VLA and introduces two coupled modules: Dual Action Experts and Expert Queries.
    Dual Action Experts keep the original AE as a frozen Prior Expert and train an Adaptation Expert for downstream action generation.
    Expert Queries capture scene and motor priors from pretrained forward paths and integrate them into the Adaptation Expert; the Prior Expert serves only as a read-only prior source, while the Adaptation Expert drives denoising and produces the final action chunk.
    }
    \label{fig:method}
\vspace{-3mm}
\end{figure}

\subsection{Overview}
\label{sec:overview}

As shown in Fig.~\ref{fig:method}, PriorVLA instantiates prior-preserving adaptation with two coupled modules: \emph{Dual Action Experts} (DAE) and \emph{Expert Queries} (EQ).
DAE separates prior preservation from task specialization by retaining the pretrained AE as a frozen \emph{Prior Expert} and training an \emph{Adaptation Expert} for downstream action generation.
EQ provides learnable interfaces for using preserved priors: \emph{Scene Queries} and \emph{Motor Queries} capture scene and motor priors from pretrained forward paths, while \emph{Action Queries} integrate these priors inside the \emph{Adaptation Expert}.
Only the \emph{Adaptation Expert}'s denoising prediction is used for action generation.

\subsection{Dual Action Experts}
\label{sec:dae}

A pretrained AE contains reusable motor priors encoded in its denoising dynamics, but full fine-tuning can shift these priors toward narrow training-distribution action patterns. PriorVLA decouples prior preservation from downstream specialization by branching the pretrained AE into \emph{Dual Action Experts}.

Specifically, we retain the original pretrained AE as a frozen \emph{Prior Expert} and introduce a trainable \emph{Adaptation Expert}, initialized from the same pretrained weights, for downstream action generation. During denoising, both experts are executed along the same noisy action trajectory. At denoising step $\tau$, they receive the same noisy action chunk $\tilde{\mathbf{a}}^{\tau}_{t:t+H-1}$. The \emph{Prior Expert} provides internal denoising representations as motor-prior features, while its denoising output is discarded. Only the denoising output of the \emph{Adaptation Expert} updates the trajectory, which produces the final action chunk after denoising. For clarity, we omit the action-horizon subscript in the recurrence:
\begin{equation}
    \tilde{\mathbf{a}}_{\mathrm{PE}}^{\tau}
    =
    \tilde{\mathbf{a}}_{\mathrm{Ada}}^{\tau}
    =
    \tilde{\mathbf{a}}^{\tau},
    \qquad
    \tilde{\mathbf{a}}^{\tau+1}
    =
    \mathrm{FM}\!\left(\tilde{\mathbf{a}}^{\tau}, f_{\mathrm{Ada}}^{\tau}\right),
\end{equation}
where $f_{\mathrm{Ada}}^{\tau}$ denotes the denoising output of the \emph{Adaptation Expert}, and $\mathrm{FM}(\cdot)$ denotes the flow-matching update. The updated chunk $\tilde{\mathbf{a}}^{\tau+1}$ is then used as the input to both experts at the next denoising step.

The \emph{Prior Expert} therefore serves as a read-only source of motor priors: its frozen weights preserve pretrained motor knowledge, while its internal denoising representations provide motor-prior features to the adapted policy, as described in Sec.~\ref{sec:queries}.

\subsection{Expert Queries}
\label{sec:queries}

Structural separation preserves the pretrained AE, but does not by itself make its priors usable for downstream action generation. Since pretrained priors are distributed across layers and denoising steps, PriorVLA introduces \emph{Expert Queries} as learnable token interfaces attached to different functional paths. They comprise three groups: \emph{Scene Queries} capture task-relevant scene priors from the VLM, \emph{Motor Queries} capture motor priors from the frozen \emph{Prior Expert}'s denoising representations, and \emph{Action Queries} integrate the captured priors inside the \emph{Adaptation Expert} for action generation.

\begin{figure}[t]
    \centering
    \includegraphics[width=1\linewidth, trim={55pt 150pt 55pt 120pt}, clip]{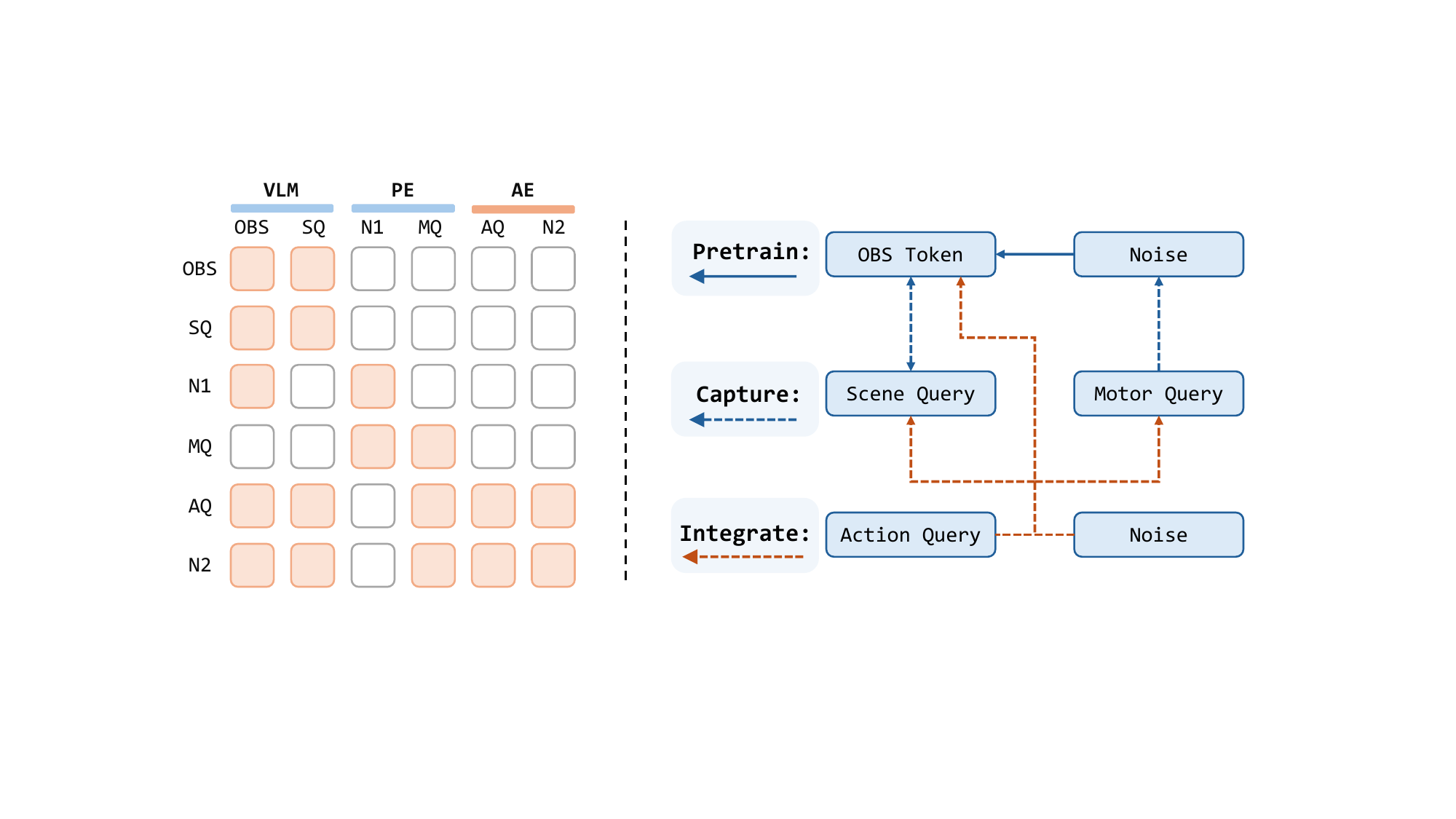}
    \caption{
    \textbf{Attention design of PriorVLA.}
    (\textbf{Left}) Attention mask over token groups in the VLM, Prior Expert (PE), and Adaptation Expert (AE). Orange cells indicate allowed attention and blank cells indicate blocked attention. OBS denotes original VLM input tokens, SQ Scene Queries, N1 PE noisy action tokens, MQ Motor Queries, AQ Action Queries, and N2 AE noisy action tokens.
    (\textbf{Right}) Information flow induced by the mask. The original VLM and PE attention paths are preserved; SQ and MQ capture scene and motor priors from pretrained forward paths, and AQ integrates them inside the AE for action generation.
    }
    \label{fig:attention}
\vspace{-4mm}
\end{figure}

We implement these interfaces through masked attention over token groups. For any token group $x$ at layer $l$, let $\mathbf{h}_x^l$ denote its hidden states, with $\mathbf{Q}_x^l=\mathbf{h}_x^l\mathbf{W}_Q^l$, $\mathbf{K}_x^l=\mathbf{h}_x^l\mathbf{W}_K^l$, and $\mathbf{V}_x^l=\mathbf{h}_x^l\mathbf{W}_V^l$. Fig.~\ref{fig:attention} summarizes the resulting attention mask and information flow; we then detail the three query groups and their roles in the adaptation path below.

\paragraph{Scene Queries.}
\emph{Scene Queries} are learnable tokens inserted into the VLM input sequence alongside the original VLM input tokens. They participate in VLM self-attention and capture task-relevant scene priors from VLM representations:
\begin{equation}
    \mathbf h_{sq}^{\,l+1}
    =
    \mathrm{Attn}\!\left(
    \mathbf Q_{sq}^l,\;
    \mathbf K_{obs}^l \| \mathbf K_{sq}^l,\;
    \mathbf V_{obs}^l \| \mathbf V_{sq}^l
    \right).
\end{equation}
Their layer-wise key-value caches provide a compact readout of visual, language, and state context, complementing the \emph{Adaptation Expert}'s direct access to OBS.

\paragraph{Motor Queries.}
\emph{Motor Queries} are learnable tokens appended to the frozen \emph{Prior Expert} to capture motor priors from its denoising representations. To preserve the pretrained denoising path, the \emph{Prior Expert}'s noisy action tokens do not attend to \emph{Scene Queries} or \emph{Motor Queries}:
\begin{equation}
    \mathbf h_{\tilde a^{pe}}^{\,l+1}
    =
    \mathrm{Attn}\!\left(
    \mathbf Q_{\tilde a^{pe}}^l,\;
    \mathbf K_{obs}^l \| \mathbf K_{\tilde a^{pe}}^l,\;
    \mathbf V_{obs}^l \| \mathbf V_{\tilde a^{pe}}^l
    \right).
\end{equation}
\emph{Motor Queries} self-attend and read only from these noisy action tokens:
\begin{equation}
    \mathbf h_{mq}^{\,l+1}
    =
    \mathrm{Attn}\!\left(
    \mathbf Q_{mq}^l,\;
    \mathbf K_{mq}^l \| \mathbf K_{\tilde a^{pe}}^l,\;
    \mathbf V_{mq}^l \| \mathbf V_{\tilde a^{pe}}^l
    \right).
\end{equation}
This one-way design preserves the frozen denoising process while exposing motor priors through the layer-wise key-value caches of \emph{Motor Queries}.

\paragraph{Action Queries.}
\emph{Action Queries} are learnable tokens inserted alongside the noisy action tokens of the \emph{Adaptation Expert}. They integrate scene priors from \emph{Scene Queries} and motor priors from \emph{Motor Queries} inside the \emph{Adaptation Expert} for downstream action generation:
\begin{equation}
    \mathbf h_{aq,\tilde a^{ae}}^{\,l+1}
    =
    \mathrm{Attn}\!\left(
    \mathbf Q_{aq,\tilde a^{ae}}^l,\;
    \mathbf K_{aq,\tilde a^{ae}}^l \| \mathbf K_{obs}^l \| \mathbf K_{sq}^l \| \mathbf K_{mq}^l,\;
    \mathbf V_{aq,\tilde a^{ae}}^l \| \mathbf V_{obs}^l \| \mathbf V_{sq}^l \| \mathbf V_{mq}^l
    \right).
\end{equation}
The raw noisy-action key-value caches of the \emph{Prior Expert} are excluded, so motor information is routed through the \emph{Motor Query} interface rather than directly copied from frozen action states. Since \emph{Action Queries} are not decoded as actions, they specialize in organizing multi-source priors inside the \emph{Adaptation Expert}.
\vspace{-1mm}

\begin{figure}[!t]
    \centering
    \includegraphics[width=1\linewidth, trim={80pt 48pt 125pt 59pt}, clip]{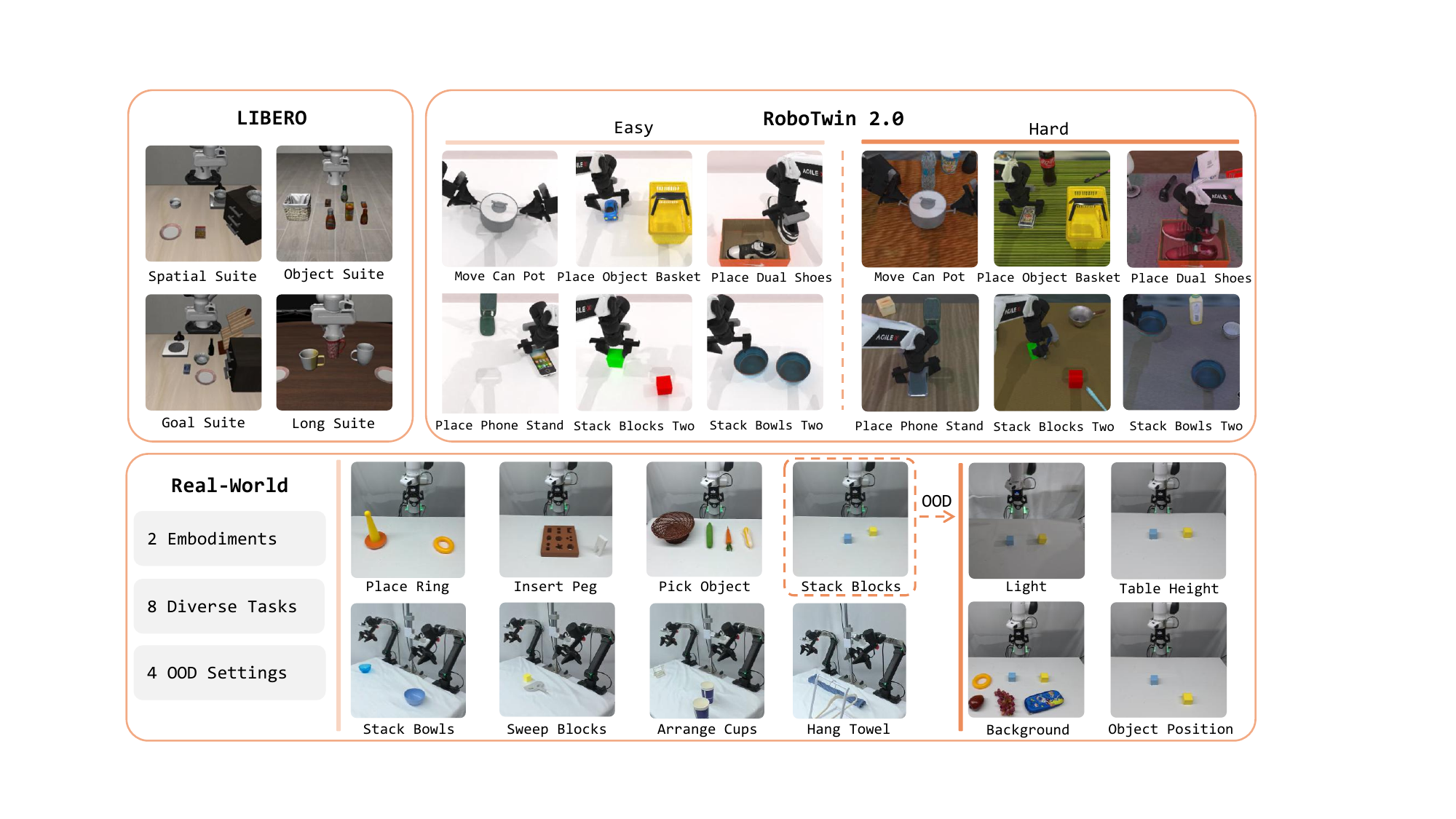}
    \caption{
    \textbf{Experimental overview.}
    We evaluate PriorVLA on RoboTwin 2.0~\citep{robotwin2}, LIBERO~\citep{libero}, and two real-world robot embodiments, covering ID/OOD generalization, data regimes, and component ablations.
    }
    \label{fig:exp_overview}
\vspace{-5mm}
\end{figure}

\subsection{Training}
\label{sec:training}

\paragraph{Trainable parameters.}
We optimize the full \emph{Adaptation Expert}, the three groups of \emph{Expert Queries}, and the VLM vision encoder, while keeping all other VLM parameters and the \emph{Prior Expert} frozen.
Overall, PriorVLA updates approximately 25\% of the parameters updated by full fine-tuning.
This keeps pretrained prior sources in the frozen VLM components and \emph{Prior Expert}, while allowing the vision encoder and adaptation branch to specialize to downstream data.

\paragraph{Objective.}
PriorVLA is trained with the standard flow-matching mean-squared error (MSE) objective, applied only to the denoising prediction of the \emph{Adaptation Expert}.
The \emph{Prior Expert}'s prediction is discarded at both training and inference time and is never included in the loss.
No auxiliary loss is imposed on the \emph{Expert Queries} or the \emph{Prior Expert}; all trainable components are optimized solely through the downstream action objective.

\section{Experiments}
\label{sec:experiments}
To comprehensively evaluate PriorVLA, we conduct experiments on RoboTwin 2.0~\cite{robotwin2}, LIBERO~\cite{libero}, and eight real-world tasks across two robot embodiments. We organize the evaluation around five questions:
(1) Can PriorVLA improve downstream adaptation over full fine-tuning and strong VLA baselines?
(2) Does preserving pretrained priors improve OOD generalization?
(3) Are the benefits stronger when downstream data coverage is limited?
(4) Do these gains transfer to real-world robots across embodiments and evaluation conditions?
(5) Are the gains explained by preserved pretrained priors and learnable query interfaces?
\vspace{-1mm}

\subsection{Experimental Setup}
\label{sec:setup}

\paragraph{Implementation details.}
PriorVLA is implemented on top of $\pi_{0.5}$~\cite{pi05}.
We train the Adaptation Expert, Expert Queries, and the VLM vision encoder, while keeping all other VLM parameters and the Prior Expert frozen.
Unless otherwise stated, models are trained for 30k steps.
Additional implementation details are provided in Appendix~\ref{app:implementation_notes} and Appendix~\ref{app:training_details}.

\paragraph{Benchmarks and evaluation.}
We evaluate PriorVLA on RoboTwin 2.0~\cite{robotwin2}, LIBERO~\cite{libero}, and real-world tasks, as summarized in Fig.~\ref{fig:exp_overview}.
RoboTwin 2.0 evaluates bimanual manipulation under Easy (ID) and Hard (OOD) modes; our standard setting uses 13 tasks with 50 clean demonstrations per task and 300 evaluation trials per setting.
We further study data scaling on the same tasks under few-shot, standard, and large-data regimes.
LIBERO evaluates four Franka manipulation suites: Spatial, Object, Goal, and Long, each with 10 tasks.
Our real-world evaluation covers eight tasks across two embodiments, a single-arm Franka robot and a dual-arm AC-One platform, with both ID/OOD evaluation and few-shot training.
OOD trials perturb four factors: Light, Background, Object Position, and Table Height.
Detailed data and evaluation protocols are provided in Appendix~\ref{app:training_data} and Appendix~\ref{app:evaluation_protocol}.

\begin{table*}[t]
\centering
\setlength{\belowcaptionskip}{7pt}
\small
\setlength{\tabcolsep}{4pt}
\renewcommand{\arraystretch}{1.04}
\caption{\textbf{RoboTwin 2.0 simulation benchmark results.} Per-task success rates (\%) on 13 tasks under standard training. Easy: in-distribution; Hard: out-of-distribution generalization. Best results are in \textbf{bold}.}
\label{tab:robotwin_sim}
\resizebox{\linewidth}{!}{
\begin{tabular}{lcccccccc>{\columncolor{gray!12}}c>{\columncolor{gray!12}}c}
\toprule
\textbf{Simulation Task}
& \multicolumn{2}{c}{\textbf{DP}}
& \multicolumn{2}{c}{\textbf{RDT}}
& \multicolumn{2}{c}{\textbf{$\pi_0$}}
& \multicolumn{2}{c}{\textbf{$\pi_{0.5}$}}
& \multicolumn{2}{c}{\textbf{PriorVLA (Ours)}} \\
\cmidrule(lr){2-3}
\cmidrule(lr){4-5}
\cmidrule(lr){6-7}
\cmidrule(lr){8-9}
\cmidrule(lr){10-11}
& \textbf{Easy} & \textbf{Hard}
& \textbf{Easy} & \textbf{Hard}
& \textbf{Easy} & \textbf{Hard}
& \textbf{Easy} & \textbf{Hard}
& \textbf{Easy} & \textbf{Hard} \\
\midrule

Grab Roller         & \textbf{98} & 0  & 74 & 43 & 96 & 80 & 97 & \textbf{93} & \textbf{98} & \textbf{93} \\
Handover Mic        & 53 & 0  & 90 & 31 & \textbf{98} & 13 & 97 & 62 & \textbf{98} & \textbf{84} \\
Lift Pot            & 39 & 0  & 72 & 9  & 84 & 36 & 67 & 25 & \textbf{96} & \textbf{66} \\
Move Can Pot        & 39 & 0  & 25 & 12 & 58 & 21 & \textbf{61} & 36 & \textbf{61} & \textbf{57} \\
Open Laptop         & 49 & 0  & 59 & 32 & 85 & 46 & \textbf{91} & 69 & \textbf{91} & \textbf{83} \\
Pick Dual Bottles   & 24 & 0  & 42 & 13 & 57 & 12 & 55 & 17 & \textbf{75} & \textbf{26} \\
Place Object Basket & 15 & 0  & 33 & 17 & 16 & 2  & 62 & 38 & \textbf{73} & \textbf{42} \\
Place Dual Shoes    & 8  & 0  & 4  & 4  & 15 & 0  & 40 & 18 & \textbf{45} & \textbf{20} \\
Place Phone Stand   & 13 & 0  & 15 & 6  & 35 & 7  & 41 & 14 & \textbf{65} & \textbf{35} \\
Put Bottles Dustbin & 22 & 0  & 21 & 4  & 54 & 13 & 60 & 43 & \textbf{64} & \textbf{45} \\
Put Object Cabinet  & 42 & 0  & 33 & 18 & 68 & 18 & 66 & \textbf{53} & \textbf{73} & 45 \\
Stack Blocks Two    & 7  & 0  & 21 & 2  & 42 & 1  & 53 & \textbf{24} & \textbf{70} & 17 \\
Stack Bowls Two     & 61 & 0  & 76 & 30 & \textbf{91} & 41 & 80 & 57 & 89 & \textbf{73} \\
\midrule
\textbf{Average}    & 36 & 0  & 44 & 17 & 62 & 22 & 67 & 42 & \textbf{77} \textcolor{teal}{(+10)} & \textbf{53} \textcolor{teal}{(+11)} \\
\bottomrule
\end{tabular}
}
\vspace{-1mm}
\end{table*}
\subsection{Experimental Results}

\noindent\textbf{Simulated Evaluation on RoboTwin 2.0.}
\label{sec:robotwin_results}
\begin{table}[t]
\centering
\setlength{\belowcaptionskip}{5pt}
\caption{\textbf{Effect of training data scale on RoboTwin 2.0.}
Average success rates (\%) over 13 tasks across data regimes. Easy/Hard denote ID/OOD evaluation, and gains over $\pi_{0.5}$ are shown in parentheses.}
\label{tab:data_regime}
\begin{tabular}{l|ccc|ccc}
\toprule
\multirow{2}{*}{Method} & \multicolumn{3}{c|}{Easy} & \multicolumn{3}{c}{Hard} \\
\cmidrule(lr){2-4}\cmidrule(lr){5-7}
 & Few & Standard & Large & Few & Standard & Large \\
\midrule
$\pi_{0.5}$~\cite{pi05} 
& 29\% & 67\% & \textbf{89\%} 
& 20\% & 42\% & 59\% \\
\rowcolor{gray!12}
PriorVLA 
& \textbf{41\%} (\textcolor{teal}{+12}) 
& \textbf{77\%} (\textcolor{teal}{+10}) 
& 88\% (-1)
& \textbf{31\%} (\textcolor{teal}{+11}) 
& \textbf{53\%} (\textcolor{teal}{+11}) 
& \textbf{65\%} (\textcolor{teal}{+6}) \\
\bottomrule
\end{tabular}
\vspace{-5.2mm}
\end{table}
As shown in Tab.~\ref{tab:robotwin_sim}, PriorVLA achieves \textbf{77\%} Easy and \textbf{53\%} Hard success under the standard setting, improving over $\pi_{0.5}$~\cite{pi05} by \textcolor{teal}{+10} and \textcolor{teal}{+11} points, respectively.
The gain is most pronounced under Hard OOD evaluation, where PriorVLA improves over $\pi_{0.5}$~\cite{pi05} on most tasks and substantially outperforms prior methods such as Diffusion Policy~\cite{diffusion_policy}, RDT~\cite{rdt}, and $\pi_0$~\cite{pi0}.
As shown in Appendix~\ref{app:training_seeds}, PriorVLA outperforms FFT under all three training seeds on a six-task subset.

We further evaluate the same 13 tasks under few-shot, standard, and large-data regimes.
As shown in Tab.~\ref{tab:data_regime}, PriorVLA consistently improves Hard performance over $\pi_{0.5}$~\cite{pi05} across all data scales, with gains of \textcolor{teal}{+11}, \textcolor{teal}{+11}, and \textcolor{teal}{+6} points.
Easy performance improves under few-shot and standard data, while becoming comparable in the large-data regime; the OOD advantage remains across all regimes.
This suggests that preserved priors are most beneficial when downstream coverage is limited, and remain useful for OOD generalization as data scales.

\noindent\textbf{Simulated Evaluation on LIBERO.}
\label{sec:libero_results}
As shown in Tab.~\ref{tab:libero}, PriorVLA achieves \textbf{99.1\%} average success across four LIBERO suites, outperforming strong VLA baselines including OpenVLA-OFT~\cite{openvla_oft} (97.1\%) and $\pi_{0.5}$~\cite{pi05} (96.9\%).
These results show that prior-preserving adaptation also improves standard in-distribution benchmark performance, even on a highly saturated benchmark.

\newsavebox{\liberoBenchmarkBox}
\newlength{\liberoBenchmarkWidth}
\begin{table*}[t]
    \centering
\setlength{\belowcaptionskip}{7pt}
    \caption{\textbf{LIBERO benchmark results.} Average success rates (\%) across four task suites. Best results in \textbf{bold}.}
    \label{tab:libero}
    \normalsize
    \renewcommand{\arraystretch}{0.98}
    \setlength{\tabcolsep}{6.5pt}
    \sbox{\liberoBenchmarkBox}{%
    \begin{tabular}{lcccc|c}
        \toprule
        Methods & Spatial & Object & Goal & Long & Avg. Success \\
        \midrule

        Diffusion Policy~\cite{diffusion_policy} &
        78.3 & 92.5 & 68.3 & 50.5 & 72.4 \\

        $\pi_{0}$-FAST~\cite{pi0_fast} &
        96.4 & 96.8 & 88.6 & 60.2 & 85.5 \\

        DreamVLA~\cite{dreamvla} &
        97.5 & 94.0 & 89.5 & 89.5 & 92.6 \\

        GR00T-N1~\cite{groot_n1} &
        94.4 & 97.6 & 93.0 & 90.6 & 93.9 \\

        $\pi_{0}$~\cite{pi0} &
        96.8 & 98.8 & 95.8 & 85.2 & 94.1 \\

        UniVLA~\cite{univla_latent} &
        95.4 & 98.8 & 93.6 & 94.0 & 95.5 \\

        F1~\cite{f1} &
        98.2 & 97.8 & 95.4 & 91.3 & 95.7 \\

        DD-VLA~\cite{dd_vla} &
        97.2 & 98.6 & 97.4 & 92.0 & 96.3 \\

        GE-Act~\cite{genie_envisioner} &
        98.2 & 97.6 & 95.8 & 94.4 & 96.5 \\

        MemoryVLA~\cite{memoryvla} &
        98.4 & 98.4 & 96.4 & 93.4 & 96.7 \\

        $\pi_{0.5}$~\cite{pi05} &
        98.8 & 98.2 & 98.0 & 92.4 & 96.9 \\

        OpenVLA-OFT~\cite{openvla_oft} &
        97.6 & 98.4 & 97.9 & 94.5 & 97.1 \\

        \midrule
        \rowcolor{gray!12}
        \textbf{PriorVLA (Ours)} &
        \textbf{99.4} & \textbf{99.8} & \textbf{99.4} & \textbf{97.6} & \textbf{99.1} \\

        \bottomrule
    \end{tabular}%
    }
    \global\liberoBenchmarkWidth=\wd\liberoBenchmarkBox
    \usebox{\liberoBenchmarkBox}%
    \vspace{-1mm}
\end{table*}

\paragraph{Zero-shot task transfer.}
As shown in Tab.~\ref{tab:zero_shot_transfer}, PriorVLA achieves \textbf{19.07\%} average success under zero-shot task transfer, outperforming FFT (8.80\%), LoRA (8.73\%), and RETAIN (9.33\%). As described in Appendix~\ref{app:zero_shot_transfer}, all methods are trained on the 10 LIBERO-10 tasks and evaluated on the 30 tasks across LIBERO-Spatial, LIBERO-Object, and LIBERO-Goal without further parameter updates. PriorVLA also achieves non-zero success on \textbf{12 of 30} target tasks, compared with at most seven for these baselines. These results show that prior-preserving adaptation improves generalization beyond the downstream training tasks.
\begin{table}[!t]
\centering
\caption{\textbf{LIBERO zero-shot transfer results.} Models are trained on LIBERO-10 and evaluated on Spatial, Object, and Goal. Average success rates (\%) are reported. Coverage counts tasks with non-zero success.}
\label{tab:zero_shot_transfer}
\normalsize
\renewcommand{\arraystretch}{0.98}
\newsavebox{\liberoTransferBox}
\newcommand{\liberoTransferRows}{%
\toprule
Methods & Spatial & Object & Goal & Avg. Success & Coverage \\
\midrule
FFT & 5.8 & \textbf{17.8} & 2.8 & 8.80 & 6/30 \\
LoRA~\cite{lora} & 10.4 & 4.0 & 11.8 & 8.73 & 7/30 \\
RETAIN~\cite{retain} & 8.8 & 17.0 & 2.2 & 9.33 & 7/30 \\
\rowcolor{gray!12} \textbf{PriorVLA (Ours)} & \textbf{21.0} & 16.6 & \textbf{19.6} & \textbf{19.07} & \textbf{12/30} \\
\bottomrule
}
\setlength{\tabcolsep}{0pt}
\sbox{\liberoTransferBox}{\begin{tabular}{lccc|cc}\liberoTransferRows\end{tabular}}
\setlength{\tabcolsep}{\dimexpr(\liberoBenchmarkWidth-\wd\liberoTransferBox)/12\relax}
\begin{tabular}{lccc|cc}
\liberoTransferRows
\end{tabular}
\end{table}

\noindent\textbf{Real-World Robot Experiments.}
\label{sec:real_robot}
We evaluate whether PriorVLA transfers to real-world tasks across two embodiments. Standard-data training uses 100--300 demonstrations depending on task difficulty, while few-shot training uses only 10 demonstrations per task.

As shown in Tabs.~\ref{tab:real_robot_standard} and~\ref{tab:real_robot_fewshot}, PriorVLA achieves the best real-world performance under both standard-data and few-shot settings.
With standard data, it reaches \textbf{81\%} ID and \textbf{57\%} OOD success, improving over $\pi_{0.5}$~\cite{pi05} by \textcolor{teal}{+12} and \textcolor{teal}{+16} points.
With only 10 demonstrations per task, PriorVLA improves ID success from 24\% to \textbf{48\%} and OOD success from 10\% to \textbf{32\%}.
These results show that PriorVLA transfers to real robots and provides larger gains when downstream demonstrations are scarce or evaluation conditions shift.
\vspace{-2mm}

\begin{table}[t]
\centering
\caption{
\textbf{Real-robot standard-data results.}
Success rates (\%) on eight real-world tasks under ID and OOD evaluation.
Average gains over $\pi_{0.5}$ are shown in parentheses.
}
\label{tab:real_robot_standard}

\setlength{\tabcolsep}{3.2pt}
\renewcommand{\arraystretch}{1.18}
\begin{tabular}{ll|cccccccc|c}
\toprule
Eval. & Method
& \begin{tabular}[c]{@{}c@{}}Place\\Ring\end{tabular}
& \begin{tabular}[c]{@{}c@{}}Insert\\Peg\end{tabular}
& \begin{tabular}[c]{@{}c@{}}Pick\\Object\end{tabular}
& \begin{tabular}[c]{@{}c@{}}Stack\\Blocks\end{tabular}
& \begin{tabular}[c]{@{}c@{}}Stack\\Bowls\end{tabular}
& \begin{tabular}[c]{@{}c@{}}Sweep\\Blocks\end{tabular}
& \begin{tabular}[c]{@{}c@{}}Arrange\\Cups\end{tabular}
& \begin{tabular}[c]{@{}c@{}}Hang\\Towel\end{tabular}
& \begin{tabular}[c]{@{}c@{}}Avg.\\Success\end{tabular} \\
\midrule

\multirow{3}{*}{ID}
& $\pi_{0.5}$~\cite{pi05} & 85 & 45 & 90 & 80 & 90 & \textbf{100} & 40 & 25 & 69 \\
& GR00T-N1.7~\cite{groot_n17} & 60 & 20 & 75 & 40 & \textbf{95} & 95 & 30 & 10 & 53 \\
& \cellcolor{gray!12}PriorVLA
& \cellcolor{gray!12}\textbf{90}
& \cellcolor{gray!12}\textbf{75}
& \cellcolor{gray!12}\textbf{90}
& \cellcolor{gray!12}\textbf{90}
& \cellcolor{gray!12}\textbf{95}
& \cellcolor{gray!12}95
& \cellcolor{gray!12}\textbf{50}
& \cellcolor{gray!12}\textbf{65}
& \cellcolor{gray!12}\textbf{81} \textcolor{teal}{(+12)} \\

\midrule

\multirow{3}{*}{OOD}
& $\pi_{0.5}$~\cite{pi05} & 45 & 10 & 55 & 50 & 70 & 75 & 15 & 10 & 41 \\
& GR00T-N1.7~\cite{groot_n17} & 15 & 0 & 40 & 20 & 80 & \textbf{85} & 5 & 0 & 31 \\
& \cellcolor{gray!12}PriorVLA
& \cellcolor{gray!12}\textbf{55}
& \cellcolor{gray!12}\textbf{30}
& \cellcolor{gray!12}\textbf{60}
& \cellcolor{gray!12}\textbf{65}
& \cellcolor{gray!12}\textbf{85}
& \cellcolor{gray!12}\textbf{85}
& \cellcolor{gray!12}\textbf{25}
& \cellcolor{gray!12}\textbf{50}
& \cellcolor{gray!12}\textbf{57} \textcolor{teal}{(+16)} \\

\bottomrule
\end{tabular}
\vspace{2mm}
\end{table}
\newlength{\realRobotMethodWidth}
\begin{table}[t]
\centering
\caption{
\textbf{Real-robot few-shot results.}
Success rates (\%) when each task is trained with only 10 demonstrations.
Average gains over $\pi_{0.5}$ are shown in parentheses.
}
\label{tab:real_robot_fewshot}

\setlength{\tabcolsep}{3.2pt}
\renewcommand{\arraystretch}{1.18}
\settowidth{\realRobotMethodWidth}{GR00T-N1.7~\cite{groot_n17}}
\makebox[\linewidth][l]{%
\begin{tabular}{ll|cccccccc|c}
\toprule
Eval. & \makebox[\realRobotMethodWidth][l]{Method}
& \begin{tabular}[c]{@{}c@{}}Place\\Ring\end{tabular}
& \begin{tabular}[c]{@{}c@{}}Insert\\Peg\end{tabular}
& \begin{tabular}[c]{@{}c@{}}Pick\\Object\end{tabular}
& \begin{tabular}[c]{@{}c@{}}Stack\\Blocks\end{tabular}
& \begin{tabular}[c]{@{}c@{}}Stack\\Bowls\end{tabular}
& \begin{tabular}[c]{@{}c@{}}Sweep\\Blocks\end{tabular}
& \begin{tabular}[c]{@{}c@{}}Arrange\\Cups\end{tabular}
& \begin{tabular}[c]{@{}c@{}}Hang\\Towel\end{tabular}
& \begin{tabular}[c]{@{}c@{}}Avg.\\Success\end{tabular} \\
\midrule

\multirow{2}{*}{ID}
& $\pi_{0.5}$~\cite{pi05} & 30 & 5 & 70 & 15 & 30 & 20 & 10 & 15 & 24 \\
& \cellcolor{gray!12}PriorVLA
& \cellcolor{gray!12}\textbf{75}
& \cellcolor{gray!12}\textbf{30}
& \cellcolor{gray!12}\textbf{85}
& \cellcolor{gray!12}\textbf{65}
& \cellcolor{gray!12}\textbf{55}
& \cellcolor{gray!12}\textbf{35}
& \cellcolor{gray!12}\textbf{15}
& \cellcolor{gray!12}\textbf{25}
& \cellcolor{gray!12}\textbf{48} \textcolor{teal}{(+24)} \\

\midrule

\multirow{2}{*}{OOD}
& $\pi_{0.5}$~\cite{pi05} & 0 & 0 & 40 & 5 & 15 & 20 & 0 & 0 & 10 \\
& \cellcolor{gray!12}PriorVLA
& \cellcolor{gray!12}\textbf{40}
& \cellcolor{gray!12}\textbf{20}
& \cellcolor{gray!12}\textbf{50}
& \cellcolor{gray!12}\textbf{55}
& \cellcolor{gray!12}\textbf{30}
& \cellcolor{gray!12}\textbf{30}
& \cellcolor{gray!12}\textbf{10}
& \cellcolor{gray!12}\textbf{20}
& \cellcolor{gray!12}\textbf{32} \textcolor{teal}{(+22)} \\

\bottomrule
\end{tabular}%
}%
\vspace{-1mm}
\end{table}


\providecommand{\cmark}{\ensuremath{\checkmark}}
\providecommand{\xmark}{\ensuremath{\times}}

\begin{table}[t]
\centering
\caption{
\textbf{Component ablations.}
Average success rates (\%) are reported on a representative subset of six RoboTwin 2.0 tasks under Easy and Hard evaluation. Params denote trainable parameters; frozen parameters are not counted. The w/o PE row removes the PE-to-AE prior pathway; because PE information is routed only through MQ, it matches the w/o MQ setting in effect.
}

\label{tab:ablation}
\vspace{-3mm}
\small
\setlength{\tabcolsep}{4.8pt}

\begin{minipage}[t]{0.48\textwidth}
\vspace{0pt}
\centering
\textbf{(a) Prior Expert ablation}\\[0.8mm]
{\renewcommand{\arraystretch}{1.32}
\begin{tabular}{l|lc|cc}
\toprule
Method & Variant & Params & Easy & Hard \\
\midrule
\multirow{4}{*}{PriorVLA}
& w/o PE       & 0.85B & 75 & 42 \\
& Random PE    & 0.85B & 75 & 43 \\
& Trainable PE & 1.28B & 73 & 44 \\
& \cellcolor{gray!12}Full
& \cellcolor{gray!12}0.85B
& \cellcolor{gray!12}\textbf{77}
& \cellcolor{gray!12}\textbf{49} \\
\bottomrule
\end{tabular}
}
\end{minipage}
\hfill
\begin{minipage}[t]{0.46\textwidth}
\vspace{0pt}
\centering
\textbf{(b) Expert Queries ablation}\\[0.8mm]
{\renewcommand{\arraystretch}{1.10}
\begin{tabular}{l|ccc|cc}
\toprule
Method & SQ & MQ & AQ & Easy & Hard \\
\midrule
\multirow{5}{*}{PriorVLA}
& \xmark & \xmark & \xmark & 61 & 28 \\
& \xmark & \cmark & \cmark & 70 & 30 \\
& \cmark & \xmark & \cmark & 75 & 42 \\
& \cmark & \cmark & \xmark & 71 & 43 \\
& \cellcolor{gray!12}\cmark
& \cellcolor{gray!12}\cmark
& \cellcolor{gray!12}\cmark
& \cellcolor{gray!12}\textbf{77}
& \cellcolor{gray!12}\textbf{49} \\
\bottomrule
\end{tabular}
}
\end{minipage}

\end{table}

\subsection{Ablation Study}
\label{sec:ablation}

We conduct component ablations on six RoboTwin 2.0 tasks and study Query configurations on four tasks, following the training and evaluation protocol in Sec.~\ref{sec:robotwin_results}. Per-task component ablation results are provided in Appendix~\ref{app:ablation_studies}.

\paragraph{Prior Expert.}
As shown in Tab.~\ref{tab:ablation}(a), PriorVLA achieves \textbf{49\%} Hard success. Removing the Prior Expert or replacing it with a random frozen branch reduces success to 42\% and 43\%, respectively, showing the benefit of pretrained motor priors. Making the Prior Expert trainable also reduces Hard success to 44\%, indicating that preserving a stable prior source is more effective than adapting both experts when generalizing beyond the downstream training distribution.

\paragraph{Expert Queries.}
As shown in Tab.~\ref{tab:ablation}(b), removing all Expert Queries reduces Hard success from \textbf{49\%} to 28\%. Among individual queries, removing Scene Queries causes the largest drop, to 30\%, while removing Motor Queries or Action Queries reduces success to 42\% and 43\%, respectively. Appendix~\ref{app:attention_analysis} further analyzes how these Queries access information during action generation.

\paragraph{Query configuration.}
As shown in Tab.~\ref{tab:query_sensitivity}, halving the Query counts reduces Hard success from \textbf{61.5\%} to 51.5\%, which remains above FFT (42.5\%). Query counts and learning rates are varied separately: the count ablation retains the 2/4/4 multipliers, while the learning rate ablation retains 64/16/64 SQ/MQ/AQ. Using the same learning rate as other trainable parameters for all Queries yields 59.3\% Hard success. These results show that retaining the full Query counts improves OOD performance, while higher Query learning rates provide a further gain.
\begin{table}[t]
\begin{minipage}[t]{0.56\linewidth}
\vspace{0pt}
\centering
\setlength{\belowcaptionskip}{6.935pt}
\caption{\textbf{Query configuration ablations.} Average success rates (\%) on four RoboTwin 2.0 tasks. LR Mult. denotes learning rate multipliers relative to the base learning rate, in SQ/MQ/AQ order.}
\label{tab:query_sensitivity}
\small
\renewcommand{\arraystretch}{1.2}
\setlength{\tabcolsep}{2.5pt}
\newcommand{\queryConfigurationRows}{%
\toprule
    Setting                                     & SQ/MQ/AQ & LR Mult. & Easy          & Hard          \\
    \midrule
    FFT                                         & --       & --       & 76.8          & 42.5          \\
    Half Query Counts                           & 32/8/32  & 2/4/4    & 88.0          & 51.5          \\
    Uniform Query LR                            & 64/16/64 & 1/1/1    & 87.5          & 59.3          \\
    \rowcolor{gray!12} \textbf{PriorVLA (Ours)} & 64/16/64 & 2/4/4    & \textbf{90.5} & \textbf{61.5} \\
    \bottomrule
}
\newsavebox{\queryConfigurationBox}
\setlength{\tabcolsep}{0pt}
\sbox{\queryConfigurationBox}{\begin{tabular}{l|cc|rr}\queryConfigurationRows\end{tabular}}
\setlength{\tabcolsep}{\dimexpr(\linewidth-\wd\queryConfigurationBox)/10\relax}
\makebox[\linewidth][l]{%
\begin{tabular}{l|cc|rr}\queryConfigurationRows\end{tabular}%
}

\end{minipage}\hfill
\begin{minipage}[t]{0.42\linewidth}
\vspace{0pt}
\centering
\caption{\textbf{Adaptation baselines.} Success rates (\%) on six RoboTwin 2.0 tasks. Best results are shown in \textbf{bold}, and second-best results are \underline{underlined}.}
\label{tab:adaptation_baselines}
\small
\renewcommand{\arraystretch}{1.0}
\setlength{\tabcolsep}{3pt}
\newcommand{\adaptationComparisonRows}{%
\toprule
Method & Easy & Hard \\
\midrule
FFT & 66.2 & 33.7 \\
LoRA~\cite{lora} & \textbf{78.7} & 26.0 \\
Frozen VLM & 35.3 & 13.7 \\
RETAIN~\cite{retain} & 61.5 & \underline{36.7} \\
\rowcolor{gray!12} \textbf{PriorVLA (Ours)} & \underline{77.3} & \textbf{48.3} \\
\bottomrule
}
\newsavebox{\adaptationComparisonBox}
\setlength{\tabcolsep}{0pt}
\sbox{\adaptationComparisonBox}{\begin{tabular}{@{\hspace{4.8pt}}>{\columncolor{white}[4.8pt][\tabcolsep]}l|rr}\adaptationComparisonRows\end{tabular}}
\setlength{\tabcolsep}{\dimexpr(\linewidth-8.1063pt-\wd\adaptationComparisonBox)/5\relax}
\makebox[\linewidth][l]{%
\begin{tabular}{@{\hspace{4.8pt}}>{\columncolor{white}[4.8pt][\tabcolsep]}l|rr}\adaptationComparisonRows\end{tabular}%
}

\end{minipage}
\end{table}

\subsection{Adaptation Baselines}
\label{sec:additional_comparisons}

As shown in Tab.~\ref{tab:adaptation_baselines}, PriorVLA achieves \textbf{48.3\%} Hard success on six RoboTwin 2.0 tasks, outperforming FFT (33.7\%), LoRA (26.0\%), and RETAIN~\cite{retain} (36.7\%). As described in Appendix~\ref{app:adaptation_baselines}, all methods share pretrained initialization, training data, and training budget. PriorVLA also maintains \textbf{77.3\%} Easy success, close to LoRA (78.7\%). These results show that PriorVLA maintains strong ID performance while achieving superior OOD generalization.

\FloatBarrier

\section{Conclusion and Limitations}
\label{sec:conclusion}
\label{sec:limitations}

In this work, we studied how to adapt pretrained Vision-Language-Action models without treating large-scale pretraining merely as initialization. We proposed PriorVLA, a prior-preserving adaptation framework that separates prior preservation from downstream specialization. Dual Action Experts retain a frozen Prior Expert as a read-only source of pretrained motor priors, while a trainable Adaptation Expert specializes to downstream action generation. Expert Queries further capture scene and motor priors from pretrained forward paths and integrate them inside the Adaptation Expert.
Across RoboTwin 2.0, LIBERO, and real-world tasks, PriorVLA improves downstream adaptation over full fine-tuning and strong VLA baselines, with the largest gains under OOD and few-shot settings. Ablations further show that these gains depend on both preserved pretrained priors and learnable query interfaces. These results highlight prior preservation as a promising principle for data-efficient and generalizable VLA adaptation.

\paragraph{Limitations.}
While PriorVLA demonstrates strong adaptation across simulation and real-world settings, several aspects remain to be further studied. Our RoboTwin 2.0 results are reported on a representative 13-task subset rather than the full 50-task benchmark because each task requires a separate model. In real-world evaluation, OOD factors are applied jointly, which reflects deployment-like shifts but does not disentangle the effect of each factor. As reported in Appendix~\ref{app:inference_cost}, PriorVLA introduces additional inference computation because the frozen Prior Expert is executed during denoising; although this overhead is manageable in our chunked-control setting, more efficient implementations or cached prior readouts could further reduce deployment cost. Finally, how scene and motor priors evolve across layers and denoising steps remains to be further studied.

\begin{ack}
This work was partly supported by the National Natural Science Foundation of China under Grant 62403012, Grant 62673073, and Zhongguancun Academy under Grant C20250503.
\end{ack}

\bibliographystyle{unsrt}
\bibliography{references}

\appendix

\clearpage
\section{Implementation Notes for PriorVLA}
\label{app:implementation_notes}

The main paper presents the architecture of PriorVLA, including Dual Action
Experts, Expert Queries, the attention design, and the training objective in
Sec.~\ref{sec:method}. This section provides implementation-oriented details
that are useful for reproducibility but omitted from the main text for clarity.

\subsection{Base VLA Backbone}
\label{app:base_vla_backbone}

PriorVLA is built on a pretrained $\pi_{0.5}$-style VLA backbone~\cite{pi05}, which couples a VLM with a flow-matching action expert. The VLM consists of a SigLIP-style
vision encoder and a Gemma-2B language backbone, and encodes multi-view images,
language prompts, and proprioceptive states. The action expert follows a
Gemma-300M-style transformer architecture and denoises future action chunks for
continuous action generation. PriorVLA keeps this pretrained backbone structure
while adding Dual Action Experts and Expert Queries for prior-preserving
adaptation.

\subsection{Dual Action Experts}
\label{app:dual_action_experts}

The Prior Expert and Adaptation Expert are both initialized from the pretrained
AE. The Prior Expert is frozen and serves as a read-only source of motor-prior
representations, while the Adaptation Expert is trainable and drives the denoising trajectory that
produces the final action chunk.

At each denoising step, both experts receive the same current noisy action
chunk. The Adaptation Expert updates the action trajectory, and the updated
chunk is used as the input to both experts at the next step. The Prior Expert is
executed along this trajectory, but its denoising output is discarded and is not
used in the loss. Its internal representations are instead exposed through Motor
Queries as motor-prior features for the Adaptation Expert.

\subsection{Expert Queries}
\label{app:query}

PriorVLA uses three groups of learnable Expert Queries. Scene Queries are
provided as additional input tokens to the VLM, alongside multi-view image
tokens, prompt tokens, and proprioceptive-state tokens. They form a compact
interface for capturing scene-prior features.

Motor Queries are provided as additional input tokens to the frozen Prior Expert
at each denoising step. They are not decoded as actions and do not update the
noisy action trajectory; instead, they read motor-prior features from the Prior
Expert forward path.

Action Queries are provided as additional input tokens to the trainable
Adaptation Expert at each denoising step. They are not decoded as actions;
instead, they integrate scene-prior features from Scene Queries and motor-prior
features from Motor Queries to guide the Adaptation Expert's denoising
prediction.

\subsection{Attention-Mask Implementation}
\label{app:attention_mask_implementation}

The attention mask is implemented with the block structure shown in
Fig.~\ref{fig:attention}. Tokens within the same block use bidirectional
self-attention, while later blocks can attend to earlier blocks but not vice
versa. This allows PriorVLA to preserve the pretrained action path of the frozen
Prior Expert while adding one-way query interfaces for prior capture.
The original VLM input tokens (OBS) and Scene Queries (SQ) belong to the same bidirectional attention block, so each can attend to the other.

For the frozen Prior Expert, noisy action tokens follow the original pretrained
action path and are not allowed to attend to Motor Queries. Motor Queries are
placed in a later readout block, so they can attend to the Prior Expert's noisy
action tokens, but they are blocked from directly attending to the VLM prefix or
Scene Queries. This makes Motor Queries focus on action-side representations
from the frozen Prior Expert. In practice, this is important because the VLM
prefix contains many more tokens than the Prior Expert action path; allowing
Motor Queries to attend to the VLM prefix can make them dominated by scene
features rather than motor-prior features.

For the trainable Adaptation Expert, Action Queries and noisy action tokens are
placed in the same adaptation block, allowing bidirectional interaction within
the Adaptation Expert. The Adaptation Expert can attend to the original VLM
features, Scene Query features, and Motor Query features, but raw noisy-action
states from the Prior Expert are excluded. Empirically, direct access to these
raw Prior Expert action states makes training less stable, while routing
motor-prior information through Motor Queries provides a compact and more stable
interface.

\section{Training Details}
\label{app:training_details}

\subsection{Training Data}
\label{app:training_data}

\paragraph{RoboTwin 2.0.}
For RoboTwin 2.0~\cite{robotwin2}, we follow the official setting and train one model per task. Given the cost of training the full
50-task benchmark, we use the 13 tasks reported in the official main table,
rather than selecting tasks based on PriorVLA performance. The standard setting
uses the official 50 clean demonstrations provided for each task. For the
few-shot and large-data settings, we use the RoboTwin 2.0 data-generation pipeline
to construct 10 and 250 clean demonstrations per task, respectively.

\paragraph{LIBERO.}
For LIBERO~\cite{libero}, we train one model for each suite: Spatial, Object, Goal, and Long.
Each suite contains 10 tasks, and LIBERO provides 50 demonstrations per task for
training.

\begin{figure}[!htbp]
    \centering
    \includegraphics[
        width=0.82\linewidth,
        trim={120pt 50pt 120pt 50pt},
        clip
    ]{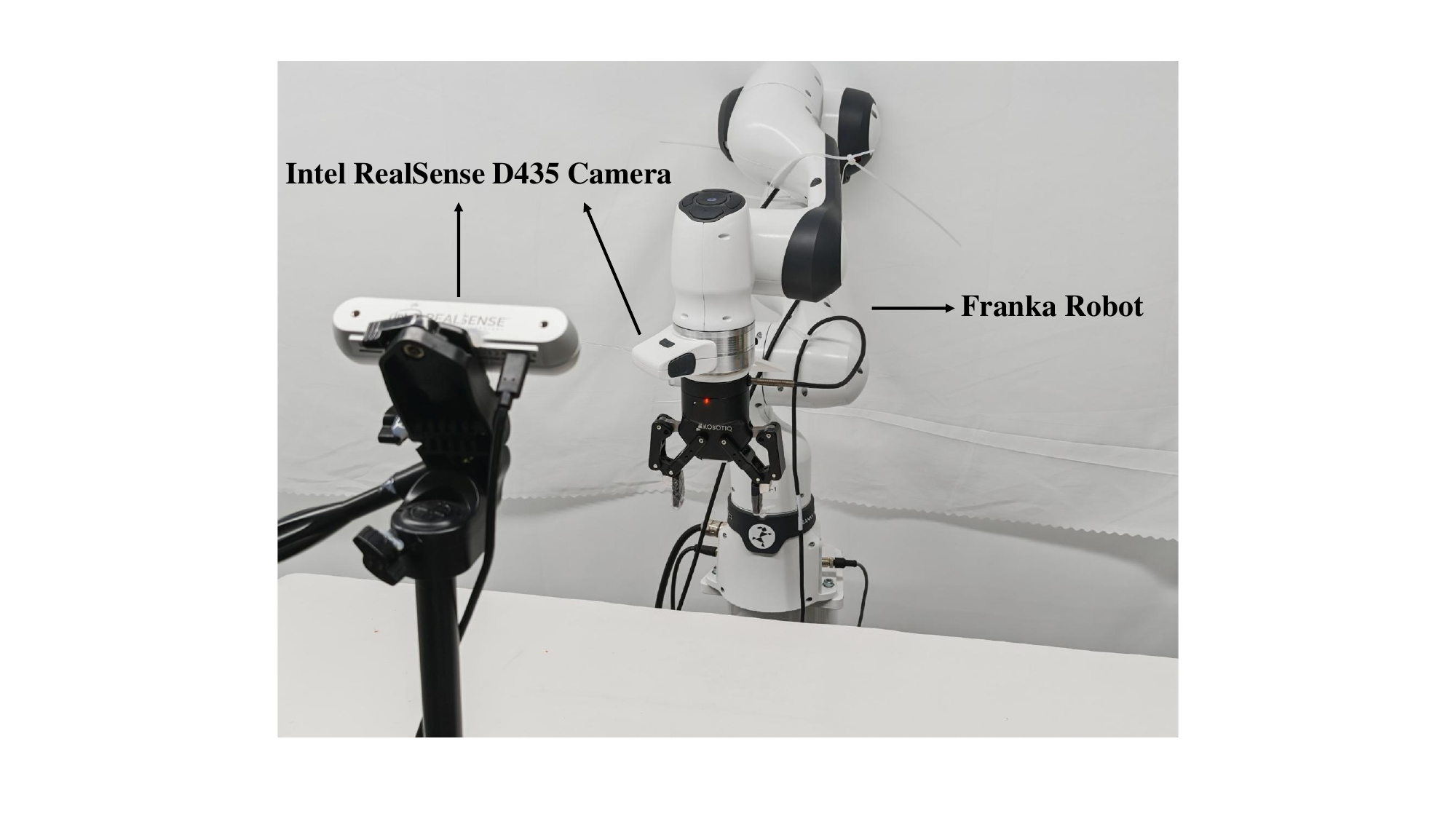}

    \vspace{2mm}

    \includegraphics[
        width=0.82\linewidth,
        trim={120pt 50pt 120pt 50pt},
        clip
    ]{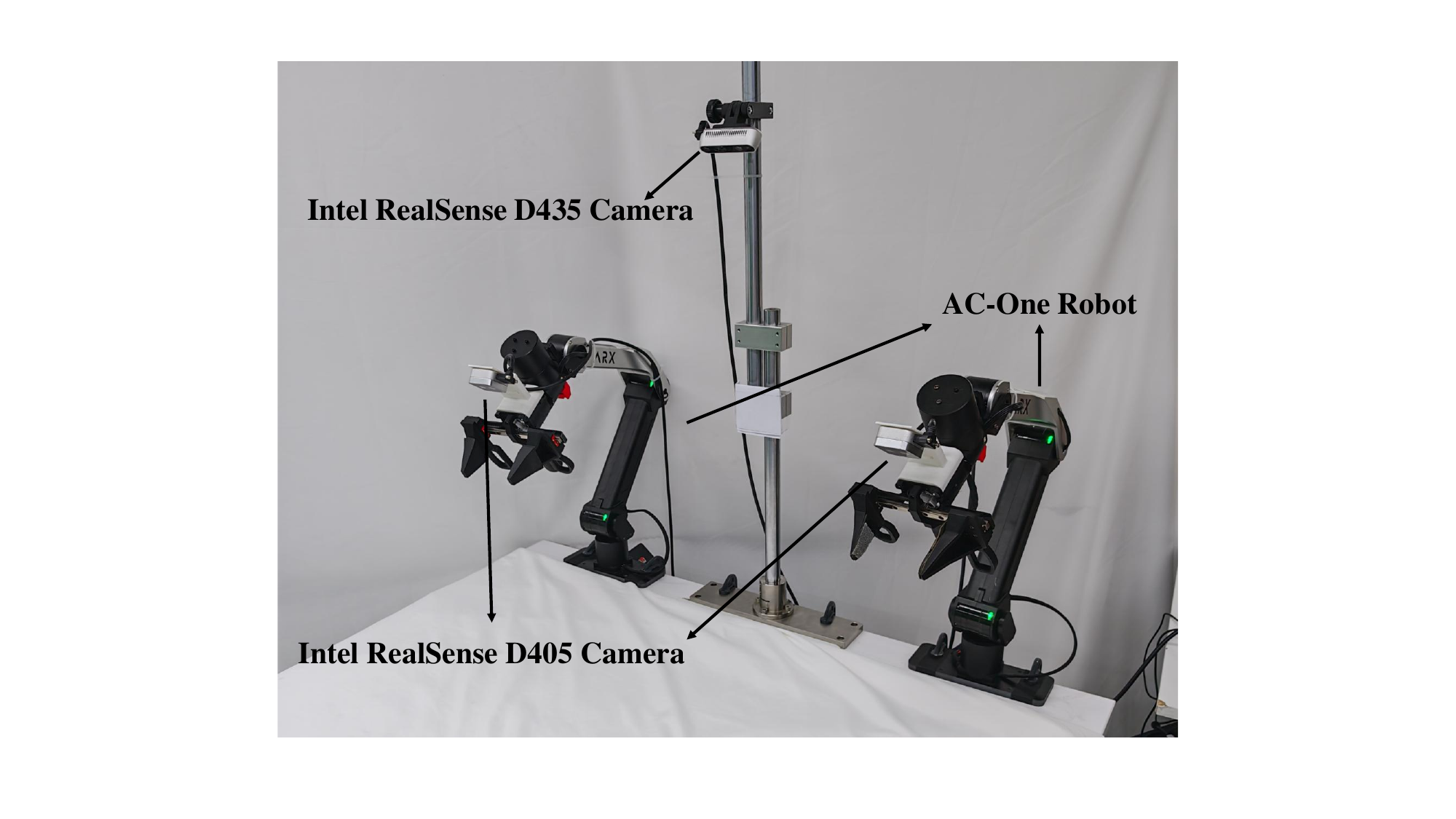}

    \caption{
    \textbf{Real-world platforms used in our experiments.}
    Top: the single-arm Franka setup with one third-person and one wrist Intel
    RealSense D435 camera. Bottom: the dual-arm AC-One setup with one top-view
    Intel RealSense D435 camera and two Intel RealSense D405 wrist cameras.
    }
    \label{fig:app_real_world_setup}
\end{figure}

\paragraph{Real-world tasks.}
We collect real-world demonstrations for eight tasks across two robot
embodiments: a single-arm Franka robot and a dual-arm AC-One platform
(Fig.~\ref{fig:app_real_world_setup}). The Franka setup uses one third-person
Intel RealSense D435 camera and one wrist Intel RealSense D435 camera. The
AC-One setup uses one top-view Intel RealSense D435 camera and two Intel
RealSense D405 wrist cameras, one for each arm. This setup covers both
single-arm and dual-arm data collection with different camera configurations.
In the standard-data setting, simple, medium, and hard tasks use 100, 200, and
300 demonstrations per task, respectively. The few-shot setting uses 10
demonstrations per task for all real-world tasks.

\begin{table}[!ht]
\centering
\small
\setlength{\tabcolsep}{7pt}
\caption{
\textbf{Real-world task suite used for training.} Demonstrations refer to the
standard-data setting; the few-shot setting uses 10 demonstrations per task.
}
\label{tab:app_real_robot_task_suite_training}
\begin{tabular}{lllc}
\toprule
\textbf{Level} & \textbf{Task} & \textbf{Robot} & \textbf{Demonstrations} \\
\midrule
Simple & Place Ring & Franka & 100 \\
Simple & Stack Blocks & Franka & 100 \\
Simple & Stack Bowls & AC-One & 100 \\
\midrule
Medium & Pick Object & Franka & 200 \\
Medium & Insert Peg & Franka & 200 \\
Medium & Sweep Blocks & AC-One & 200 \\
Medium & Arrange Cups & AC-One & 200 \\
\midrule
Hard & Hang Towel & AC-One & 300 \\
\bottomrule
\end{tabular}
\end{table}

\subsection{Hyperparameters}
\label{app:hyperparameters}

Across all experiments, we use AdamW with global-norm gradient clipping at 1.0.
Trainable parameters are optimized in float32, and frozen pretrained parameters
are stored in bfloat16. Unless otherwise specified, all runs use random seed 42.
PriorVLA applies grouped learning-rate multipliers to four parameter groups:
Scene Queries, Motor Queries, Action Queries, and the remaining trainable
parameters. The corresponding multipliers are 2.0, 4.0, 4.0, and 1.0.
We use 64 Scene Queries, 16 Motor Queries, and 64 Action Queries for RoboTwin 2.0 and real-world experiments, and 8, 2, and 8, respectively, for LIBERO.
The benchmark-specific hyperparameters are summarized in
Table~\ref{tab:app_benchmark_hparams}.

\paragraph{RoboTwin 2.0.}
For RoboTwin 2.0~\cite{robotwin2}, we train a separate model for each task for 30k steps with action horizon $H=50$. The few-shot and standard-data settings use global batch size
32, base peak learning rate $2.5\times10^{-5}$, and base decay learning rate
$2.5\times10^{-6}$. The large-data setting uses global batch size 128 and
doubles the base peak and decay learning rates to $5.0\times10^{-5}$ and
$5.0\times10^{-6}$. All RoboTwin 2.0 runs use 1k warmup steps, cosine decay over
30k steps, and EMA decay 0.99.

\paragraph{LIBERO.}
For LIBERO~\cite{libero}, we train a separate model for each suite for 30k steps with action horizon $H=10$ and global batch size 256.
To align with the standard $\pi_{0.5}$~\cite{pi05} LIBERO~\cite{libero} setting, we keep the official training schedule unchanged:
10k warmup steps, peak and decay learning rates of $5.0\times10^{-5}$, 1M decay
steps, and EMA decay 0.999.

\paragraph{Real-world tasks.}
For real-world tasks, we train one model per task for 30k steps with action
horizon $H=50$ and global batch size 32. We use the same base learning-rate
schedule as RoboTwin 2.0: 1k warmup to $2.5\times10^{-5}$, followed by cosine decay
to $2.5\times10^{-6}$ over 30k steps. The EMA decay is 0.99.

\begin{table}[!ht]
\centering
\small
\setlength{\tabcolsep}{5.5pt}
\caption{
\textbf{Benchmark-specific training hyperparameters.}
RoboTwin 2.0 values correspond to the few-shot and standard-data settings; the
large-data setting uses global batch size 128 and doubled base learning rates.
}
\label{tab:app_benchmark_hparams}
\begin{tabular}{lccc}
\toprule
\textbf{Hyperparameter} & \textbf{RoboTwin 2.0} & \textbf{LIBERO} & \textbf{Real-world} \\
\midrule
Training granularity & Per task & Per suite & Per task \\
Training steps & 30k & 30k & 30k \\
Global batch size & 32 & 256 & 32 \\
Action horizon $H$ & 50 & 10 & 50 \\
Warmup steps & 1k & 10k & 1k \\
Base peak LR & $2.5{\times}10^{-5}$ & $5.0{\times}10^{-5}$ & $2.5{\times}10^{-5}$ \\
Base decay LR & $2.5{\times}10^{-6}$ & $5.0{\times}10^{-5}$ & $2.5{\times}10^{-6}$ \\
Decay steps & 30k & 1M & 30k \\
EMA decay & 0.99 & 0.999 & 0.99 \\
\bottomrule
\end{tabular}
\end{table}

\subsection{Training Procedure}
\label{app:training_procedure}

Each training sample is constructed from a demonstration trajectory and contains
multi-view RGB observations, a language prompt, proprioceptive states, and a
future action chunk. Image frames are aligned with state-action records, and
future action chunks are constructed according to the benchmark-specific action
horizon $H$. Actions are represented as delta actions and normalized using
training-set statistics. For the dual-arm AC-One platform, we concatenate the
left- and right-arm delta actions into a joint 14-D action representation.

Images are resized to $224 \times 224$ before being fed into the policy. During
training, we use the default $\pi_{0.5}$ image preprocessing and augmentation
pipeline, and disable image augmentations during evaluation.

For RoboTwin 2.0 and real-world experiments, we report the final checkpoint after
30k training steps. LIBERO follows the standard $\pi_{0.5}$ LIBERO checkpointing
and evaluation recipe. 

\paragraph{Compute resources.}
Main experiments are trained on 8 NVIDIA GPUs. RoboTwin 2.0 experiments are trained
on H20 GPUs, while LIBERO and real-world experiments are trained on A100 GPUs.
Table~\ref{tab:app_compute_resources} reports representative wall-clock time per
trained model and the memory capacity of the GPUs used in our experiments.

\begin{table}[!ht]
\centering
\small
\setlength{\tabcolsep}{4.5pt}
\renewcommand{\arraystretch}{1.05}
\caption{
\textbf{Representative compute resources.} Wall-clock time is reported per
trained model and is approximate. Mem./GPU denotes the memory capacity of each
GPU. The $\pi_{0.5}$ LIBERO baseline is taken from the official report and was
not re-trained in our compute accounting.
}
\label{tab:app_compute_resources}
\begin{tabular}{@{}llccccc@{}}
\toprule
\textbf{Setting} & \textbf{Method} & \textbf{GPUs} & \textbf{Mem./GPU} &
\textbf{Batch} & \textbf{Steps} & \textbf{Time} \\
\midrule

\multirow{2}{*}{RoboTwin 2.0 standard}
& $\pi_{0.5}$ & 8 H20 & 96GB & 32 & 30k & $\sim$6.8h \\
& \cellcolor{gray!12}PriorVLA
& \cellcolor{gray!12}8 H20
& \cellcolor{gray!12}96GB
& \cellcolor{gray!12}32
& \cellcolor{gray!12}30k
& \cellcolor{gray!12}$\sim$5.6h \\
\midrule

\multirow{2}{*}{LIBERO}
& $\pi_{0.5}$ & -- & -- & -- & -- & -- \\
& \cellcolor{gray!12}PriorVLA
& \cellcolor{gray!12}8 A100
& \cellcolor{gray!12}80GB
& \cellcolor{gray!12}256
& \cellcolor{gray!12}30k
& \cellcolor{gray!12}$\sim$23.6h \\
\midrule

\multirow{2}{*}{Real-world tasks}
& $\pi_{0.5}$ & 8 A100 & 80GB & 32 & 30k & $\sim$6.5h \\
& \cellcolor{gray!12}PriorVLA
& \cellcolor{gray!12}8 A100
& \cellcolor{gray!12}80GB
& \cellcolor{gray!12}32
& \cellcolor{gray!12}30k
& \cellcolor{gray!12}$\sim$5.0h \\

\bottomrule
\end{tabular}
\end{table}

\section{Evaluation Protocol}
\label{app:evaluation_protocol}

This section describes the evaluation protocols used for RoboTwin 2.0, LIBERO, and
real-world experiments. We report success rate as the primary metric. RoboTwin 2.0
and LIBERO use their official success signals, while real-world experiments use
task-specific success criteria.

\subsection{RoboTwin 2.0 Evaluation}
\label{app:robotwin2_eval_protocol}

\paragraph{Protocol.}
For RoboTwin 2.0, we evaluate the same 13 tasks used for training under the
official Easy and Hard modes. Easy serves as the ID setting, and Hard serves as
the OOD setting with randomized scene factors. For each method, we evaluate the
final 30k checkpoint with 300 rollouts per task for each mode. Success is
measured by the official RoboTwin 2.0 simulator signal.

\paragraph{Easy and Hard modes.}
Easy and Hard modes use the same robot embodiment, camera setup, and
observation/state inputs, and differ only in domain randomization. The Easy mode
disables scene randomization and serves as the ID setting. The Hard mode enables
background randomization, table clutter, lighting variation, and table-height
perturbation, and serves as the OOD setting. The main differences are summarized
in Table~\ref{tab:app_robotwin2_easy_hard_config}.

\begin{table}[!ht]
\centering
\small
\setlength{\tabcolsep}{6pt}
\caption{
\textbf{RoboTwin 2.0 Easy and Hard modes.} The Easy mode is the clean ID setting,
while the Hard mode enables domain randomization and is used as the OOD setting.
}
\label{tab:app_robotwin2_easy_hard_config}
\begin{tabular}{lcc}
\toprule
\textbf{Field} & \textbf{Easy mode / ID} & \textbf{Hard mode / OOD} \\
\midrule
\multicolumn{3}{l}{\textit{Domain randomization}} \\
Background randomization & Off & On \\
Table clutter & Off & On \\
Clean background rate & 1.00 & 0.02 \\
Table-height perturbation & 0 & 0.03 \\
Lighting randomization & Off & On \\
Extreme lighting rate & 0 & 0.02 \\
\midrule
\multicolumn{3}{l}{\textit{Shared setup}} \\
Robot embodiment & Aloha-AgileX & Aloha-AgileX \\
Head / wrist camera & D435 / D435 & D435 / D435 \\
RGB observation & Yes & Yes \\
End-effector state / qpos & Yes / Yes & Yes / Yes \\
\bottomrule
\end{tabular}
\end{table}

\subsection{LIBERO Evaluation}
\label{app:libero_eval_protocol}

For LIBERO, we evaluate the four suites used in the main paper: Spatial, Object,
Goal, and Long. Each suite contains 10 tasks, and each task is evaluated with
50 rollouts, yielding 500 rollouts per suite. We use the initial states provided
by LIBERO for each task, and success is measured by the LIBERO environment success
signal. The evaluation seed is 7, and the policy replans every 10 environment
steps.

For checkpointing, we follow the standard $\pi_{0.5}$ LIBERO evaluation recipe.
Checkpoints from 20k steps onward are evaluated at fixed 1k-step intervals, and
the best checkpoint is reported.

\subsection{Real-World Evaluation}
\label{app:real_world_eval_protocol}

\paragraph{Protocol.}
For real-world evaluation, we evaluate the final 30k checkpoint of each trained
policy on 20 ID trials and 20 OOD trials per task. The same policy is evaluated
under both settings without retraining. For all real-world tasks, the policy
predicts an action chunk of length $H=50$, executes the first 15 actions, and
then re-queries the policy with a new observation. Success rates are reported
separately for ID and OOD trials.

\paragraph{ID and OOD settings.}
ID trials follow the nominal task setup used during data collection. OOD trials
use a harder test-time setup that jointly changes four dimensions: Light,
Background, Object Position, and Table Height. All four changes are applied
together in each OOD trial, while the task instruction and robot platform remain
unchanged. All compared methods are evaluated under the same ID/OOD protocol.
The OOD dimensions are summarized in Table~\ref{tab:app_real_world_ood_design}.

\begin{table}[!ht]
\centering
\small
\setlength{\tabcolsep}{5pt}
\caption{
\textbf{Real-world OOD dimensions.} Each OOD trial jointly applies all four
dimensions while keeping the task instruction and robot platform unchanged.
}
\label{tab:app_real_world_ood_design}
\begin{tabular}{p{0.24\linewidth}p{0.66\linewidth}}
\toprule
\textbf{Dimension} & \textbf{OOD Perturbation} \\
\midrule
Light & The scene is evaluated under darker lighting conditions. \\
Background & Distracting objects are added to the background or workspace. \\
Object Position & Task objects are initialized at positions different from the nominal setup. \\
Table Height & The table is raised by 2 cm relative to the nominal setup. \\
\bottomrule
\end{tabular}
\end{table}

\paragraph{Success criteria.}
A real-world trial is counted as successful only if the task-specific completion
condition is satisfied before the trial ends. A trial is counted as a failure if
the task is not completed, the object is dropped irrecoverably, a safety stop or
collision occurs, or human intervention is required. The task-specific success
criteria are listed in Table~\ref{tab:app_real_world_success_criteria}.

\begin{table}[!ht]
\centering
\small
\setlength{\tabcolsep}{5pt}
\caption{
\textbf{Task-specific success criteria for real-world evaluation.}
}
\label{tab:app_real_world_success_criteria}
\begin{tabular}{p{0.24\linewidth}p{0.66\linewidth}}
\toprule
\textbf{Task} & \textbf{Success Criterion} \\
\midrule
Place Ring & The ring is inserted onto the vertical peg. \\
Insert Peg & The peg is inserted into the matching slot. \\
Pick Object & The target object is placed into the basket after release. \\
Stack Blocks & The target block is stacked on the other block and remains stable after release. \\
Stack Bowls & The bowls are stacked together, and the upper bowl does not touch the table. \\
Sweep Blocks & The blocks are pushed into the container and do not remain on the table. \\
Arrange Cups & The cups are stacked together and fully placed into the rack. \\
Hang Towel & The towel remains on the hanger after the hanger is lifted and returned. \\
\bottomrule
\end{tabular}
\end{table}

\section{Additional Results}
\label{app:additional_results}

This section reports additional experimental records that are not fully expanded in the main paper. We provide detailed RoboTwin 2.0 results and additional analysis of LIBERO zero-shot task transfer. The full per-task RoboTwin 2.0 results across few-shot, standard, and large-data settings are reported in Table~\ref{tab:app_rt2_full_per_task}. The full per-task ablation results on the six-task RoboTwin 2.0 subset are reported in Table~\ref{tab:app_ablation_pertask}.

\subsection{RoboTwin 2.0 Results across Data Regimes}
\label{app:full_robotwin2_data_regimes}

The main paper reports the average RoboTwin 2.0 data-regime comparison. Here, we provide the per-task results across few-shot, standard, and large-data settings in Table~\ref{tab:app_rt2_full_per_task}. These results show that PriorVLA improves OOD performance across data regimes. In the large-data setting, the in-distribution Easy average of PriorVLA is close to $\pi_{0.5}$, while Hard performance remains better, suggesting that additional demonstrations reduce the need for preserved priors on ID scenes but do not remove the need for preserved priors under distribution shift.

\begin{table}[!t]
\centering
\small
\caption{
Full per-task RoboTwin 2.0 results across data regimes. Success rates (\%) are reported under Easy and Hard evaluation. Best values within each data regime are shown in bold, and PriorVLA columns are shaded.
}
\label{tab:app_rt2_full_per_task}
\resizebox{\linewidth}{!}{
\begin{tabular}{lcc>{\columncolor{gray!12}}c>{\columncolor{gray!12}}ccc>{\columncolor{gray!12}}c>{\columncolor{gray!12}}ccc>{\columncolor{gray!12}}c>{\columncolor{gray!12}}c}
\toprule
\multirow{2}{*}{Task}
& \multicolumn{2}{c}{$\pi_{0.5}$ Few}
& \multicolumn{2}{c}{PriorVLA Few}
& \multicolumn{2}{c}{$\pi_{0.5}$ Standard}
& \multicolumn{2}{c}{PriorVLA Standard}
& \multicolumn{2}{c}{$\pi_{0.5}$ Large}
& \multicolumn{2}{c}{PriorVLA Large} \\
\cmidrule(lr){2-3}\cmidrule(lr){4-5}\cmidrule(lr){6-7}
\cmidrule(lr){8-9}\cmidrule(lr){10-11}\cmidrule(lr){12-13}
& Easy & Hard & Easy & Hard & Easy & Hard & Easy & Hard & Easy & Hard & Easy & Hard \\
\midrule
Grab Roller & 41 & 46 & \textbf{79} & \textbf{73} & 97 & \textbf{93} & \textbf{98} & \textbf{93} & \textbf{100} & 93 & \textbf{100} & \textbf{98} \\
Handover Mic & 85 & 60 & \textbf{88} & \textbf{77} & 97 & 62 & \textbf{98} & \textbf{84} & \textbf{99} & 65 & \textbf{99} & \textbf{83} \\
Lift Pot & 8 & 5 & \textbf{51} & \textbf{28} & 67 & 25 & \textbf{96} & \textbf{66} & 99 & 51 & \textbf{100} & \textbf{69} \\
Move Can Pot & 45 & \textbf{32} & \textbf{47} & 28 & \textbf{61} & 36 & \textbf{61} & \textbf{57} & 90 & \textbf{70} & \textbf{92} & 68 \\
Open Laptop & 69 & 37 & \textbf{78} & \textbf{60} & \textbf{91} & 69 & \textbf{91} & \textbf{83} & \textbf{98} & 75 & 96 & \textbf{95} \\
Pick Dual Bottles & 3 & 2 & \textbf{10} & \textbf{19} & 55 & 17 & \textbf{75} & \textbf{26} & 88 & 42 & \textbf{93} & \textbf{45} \\
Place Object Basket & 31 & 15 & \textbf{34} & \textbf{20} & 62 & 38 & \textbf{73} & \textbf{42} & \textbf{76} & \textbf{42} & 72 & 40 \\
Place Dual Shoes & 1 & 1 & \textbf{4} & \textbf{2} & 40 & 18 & \textbf{45} & \textbf{20} & \textbf{79} & \textbf{42} & 68 & 38 \\
Place Phone Stand & 1 & 4 & \textbf{8} & \textbf{6} & 41 & 14 & \textbf{65} & \textbf{35} & \textbf{74} & 36 & \textbf{74} & \textbf{53} \\
Put Bottles Dustbin & 16 & 8 & \textbf{21} & \textbf{18} & 60 & 43 & \textbf{64} & \textbf{45} & \textbf{83} & 65 & 79 & \textbf{66} \\
Put Object Cabinet & 25 & 17 & \textbf{32} & \textbf{21} & 66 & \textbf{53} & \textbf{73} & 45 & \textbf{91} & 67 & 86 & \textbf{70} \\
Stack Blocks Two & 8 & 4 & \textbf{13} & \textbf{5} & 53 & \textbf{24} & \textbf{70} & 17 & 87 & 34 & \textbf{91} & \textbf{43} \\
Stack Bowls Two & 43 & 27 & \textbf{67} & \textbf{43} & 80 & 57 & \textbf{89} & \textbf{73} & \textbf{94} & 82 & 89 & \textbf{84} \\
\midrule
Average & 29 & 20 & \textbf{41} & \textbf{31} & 67 & 42 & \textbf{77} & \textbf{53} & \textbf{89} & 59 & 88 & \textbf{65} \\
\bottomrule
\end{tabular}
}
\end{table}

\subsection{Consistency and Sign-Test Analysis}
\label{app:consistency_statistics}

\begin{table}[!t]
\centering
\small
\setlength{\tabcolsep}{5pt}
\caption{
\textbf{Consistency of improvements over $\pi_{0.5}$.}
We count task-setting pairs where PriorVLA improves over $\pi_{0.5}$.
Ties are excluded when computing the sign-test statistics.
Avg. Gain denotes the aggregate average success-rate difference in points.
Sign-test $p$ denotes the one-sided exact binomial $p$-value under the null
hypothesis that either method is equally likely to perform better.
}
\label{tab:app_consistency_statistics}
\begin{tabular}{lcccc}
\toprule
\textbf{Setting} & \textbf{Non-tie Pairs} & \textbf{Improved} & \textbf{Avg. Gain (pts)} & \textbf{Sign-test $p$} \\
\midrule
RoboTwin 2.0 few-shot Easy & 13 & 13 & +12 & $1.2{\times}10^{-4}$ \\
RoboTwin 2.0 few-shot Hard & 13 & 12 & +11 & $1.7{\times}10^{-3}$ \\
RoboTwin 2.0 standard Easy & 11 & 11 & +10 & $4.9{\times}10^{-4}$ \\
RoboTwin 2.0 standard Hard & 12 & 10 & +11 & $1.9{\times}10^{-2}$ \\
RoboTwin 2.0 large-data Hard & 13 & 10 & +6 & $4.6{\times}10^{-2}$ \\
Real-world standard ID/OOD & 15 & 14 & +14 & $4.9{\times}10^{-4}$ \\
Real-world few-shot ID/OOD & 16 & 16 & +23 & $1.5{\times}10^{-5}$ \\
\bottomrule
\end{tabular}
\end{table}

Beyond aggregate averages, we analyze whether PriorVLA's gains are consistent
across task-setting pairs. For RoboTwin 2.0, each pair corresponds to one task
under a specific data regime and evaluation mode. For real-world experiments,
each pair corresponds to one task under either the ID or OOD evaluation setting.
As shown in Table~\ref{tab:app_consistency_statistics}, PriorVLA improves on a
large majority of non-tie task-setting pairs. The gains are therefore not driven
by a single task or a single setting, but are consistent across RoboTwin 2.0
few-shot, standard, and OOD settings, as well as standard-data and few-shot
real-world evaluations. For real-world evaluations, each ID or OOD aggregate
contains 8 tasks with 20 trials per task, yielding 160 trials per evaluation
condition.

\subsection{RoboTwin 2.0 Ablation Results}
\label{app:ablation_studies}

We conduct ablations on a representative subset of six RoboTwin 2.0 tasks:
Handover Mic, Lift Pot, Place Dual Shoes, Place Phone Stand, Stack Blocks Two,
and Stack Bowls Two. These tasks cover grasping, placement, stacking, and
longer-horizon manipulation. The main paper reports the average ablation
comparison. Here, we provide the per-task Easy and Hard results in
Table~\ref{tab:app_ablation_pertask}, including Prior Expert controls, Expert
Query ablations, and the vision-encoder ablation.

\begin{table}[!ht]
\centering
\small
\caption{
Per-task ablation results on the six-task RoboTwin 2.0 subset. Success rates (\%) are reported under Easy and Hard evaluation. Best values in each column are shown in bold, and the PriorVLA row is shaded.
}
\label{tab:app_ablation_pertask}
\resizebox{\linewidth}{!}{
\begin{tabular}{lcccccccccccccc}
\toprule
\multirow{2}{*}{Variant}
& \multicolumn{2}{c}{Average}
& \multicolumn{2}{c}{Handover Mic}
& \multicolumn{2}{c}{Lift Pot}
& \multicolumn{2}{c}{Place Dual Shoes}
& \multicolumn{2}{c}{Place Phone Stand}
& \multicolumn{2}{c}{Stack Blocks Two}
& \multicolumn{2}{c}{Stack Bowls Two} \\
\cmidrule(lr){2-3}\cmidrule(lr){4-5}\cmidrule(lr){6-7}
\cmidrule(lr){8-9}\cmidrule(lr){10-11}\cmidrule(lr){12-13}
\cmidrule(lr){14-15}
& Easy & Hard & Easy & Hard & Easy & Hard & Easy & Hard & Easy & Hard & Easy & Hard & Easy & Hard \\
\midrule
Baseline & 63 & 33 & 97 & 62 & 67 & 25 & 40 & 18 & 41 & 14 & 53 & \textbf{24} & 80 & 57 \\
Random PE & 75 & 43 & 96 & 79 & 95 & 48 & 43 & 11 & 61 & 34 & 69 & 15 & 87 & 72 \\
Trainable PE & 73 & 44 & \textbf{98} & 76 & 94 & 58 & 34 & 10 & 59 & 34 & 66 & 14 & 88 & 74 \\
w/o SQ/MQ/AQ & 61 & 28 & 94 & 27 & 93 & 48 & 25 & 12 & 45 & 8 & 30 & 15 & 80 & 58 \\
w/o AQ & 71 & 43 & 95 & 82 & 92 & 47 & 30 & 11 & 55 & 30 & 67 & 14 & 88 & 75 \\
w/o SQ & 70 & 30 & \textbf{98} & 26 & 95 & 49 & 32 & 3 & 52 & 20 & 59 & 14 & 83 & 70 \\
w/o MQ & 75 & 42 & \textbf{98} & 80 & \textbf{96} & 37 & 40 & 14 & 60 & 34 & \textbf{70} & 15 & 86 & 71 \\
Frozen ViT & 65 & 37 & 87 & 66 & 83 & 20 & 24 & 10 & 45 & \textbf{35} & 65 & 14 & 87 & \textbf{76} \\
\rowcolor{gray!12} PriorVLA & \textbf{77} & \textbf{49} & \textbf{98} & \textbf{84} & \textbf{96} & \textbf{66} & \textbf{45} & \textbf{20} & \textbf{65} & \textbf{35} & \textbf{70} & 17 & \textbf{89} & 73 \\
\bottomrule
\end{tabular}
}
\end{table}
\paragraph{Analysis.}
The Prior Expert controls support the role of a pretrained and preserved prior
source. The w/o PE and w/o MQ variants have the same effect on action generation because PE information is routed to the Adaptation Expert only through Motor Queries. Thus, w/o PE removes the prior source, while w/o MQ keeps the frozen branch but disables its usable interface. Their similar performance confirms that the Prior Expert must be connected through a learnable Motor Query interface to benefit adaptation. Random PE reaches 75\% Easy and 43\% Hard, close to the w/o MQ variant,
indicating that a random frozen branch does not provide useful motor priors.
Trainable PE reaches 73\% Easy and 44\% Hard, also below full PriorVLA,
suggesting that updating the prior source weakens its usefulness. Removing all
Expert Queries reduces Hard performance from 49\% to 28\%, showing that keeping
a frozen Prior Expert alone is insufficient without learnable interfaces.
Among single-query ablations, removing Scene Queries causes the largest Hard
degradation, while removing Motor Queries or Action Queries also reduces OOD
performance. Freezing the vision encoder decreases both Easy and Hard
performance, showing that visual adaptation remains complementary to prior
preservation.

The Query configuration ablations, initialization comparison, and attention analysis use the same four RoboTwin 2.0 tasks: Handover Mic, Lift Pot, Stack Blocks Two, and Stack Bowls Two. These experiments use separate runs from the six-task component ablations.

On this subset, initializing the VLM from PaliGemma achieves 60.8\% Easy and 2.8\% Hard success, compared with 90.5\% and 61.5\% when initializing it from $\pi_{0.5}$ as in PriorVLA. Both settings use the same architecture, PE and AE initialization, Query counts, data, and training budget.

\subsection{Adaptation Baselines}
\label{app:adaptation_baselines}

We compare adaptation baselines on the six-task RoboTwin 2.0 subset described in
Appendix~\ref{app:ablation_studies}, using the same pretrained initialization,
training data, and training budget. Table~\ref{tab:app_adaptation_baselines}
provides the full comparison corresponding to Table~\ref{tab:adaptation_baselines},
including First-4 VLM Tuning and both RETAIN configurations.
This comparison uses separate runs from the component ablations in Table~\ref{tab:app_ablation_pertask}.

LoRA follows the default openpi configuration, with ranks of 16 and 32 for the VLM language backbone and action expert, respectively, applied to attention and feed-forward modules. Frozen VLM freezes the entire VLM, including the vision encoder, and trains only the action expert. First-4 VLM Tuning updates the first four VLM transformer layers while freezing the remaining VLM transformer layers. RETAIN interpolates pretrained and finetuned weights as $\theta = (1-\alpha)\theta_{\mathrm{pre}}+\alpha\theta_{\mathrm{ft}}$.

\begin{table}[!t]
\centering
\small
\setlength{\tabcolsep}{7pt}
\caption{
Adaptation baselines on the six-task RoboTwin 2.0 subset. Average success rates
(\%) are reported under Easy and Hard evaluation. Best values in each column
are shown in bold, and the PriorVLA row is shaded.
}
\label{tab:app_adaptation_baselines}
\begin{tabular}{lcc}
\toprule
Method & Easy & Hard \\
\midrule
FFT & 66.2 & 33.7 \\
LoRA~\cite{lora} & \textbf{78.7} & 26.0 \\
Frozen VLM & 35.3 & 13.7 \\
First-4 VLM Tuning & 70.8 & 32.4 \\
RETAIN~\cite{retain} ($\alpha = 0.95$) & 54.5 & 32.2 \\
RETAIN~\cite{retain} ($\alpha = 0.98$) & 61.5 & 36.7 \\
\rowcolor{gray!12} \textbf{PriorVLA (Ours)} & 77.3 & \textbf{48.3} \\
\bottomrule
\end{tabular}
\end{table}

\subsection{Zero-shot Task Transfer}
\label{app:zero_shot_transfer}

For the zero-shot transfer experiment in Table~\ref{tab:zero_shot_transfer}, all
methods are adapted on LIBERO-10 and evaluated on LIBERO-Spatial, LIBERO-Object,
and LIBERO-Goal, without target-suite demonstrations or further parameter
updates. RETAIN~\cite{retain} uses $\alpha = 0.9$ in this comparison.

Removing Scene Queries reduces average success from 19.07\% to 16.93\%,
supporting their contribution to generalization beyond the downstream training
tasks.

\subsection{Results across Training Seeds}
\label{app:training_seeds}

We train FFT and PriorVLA with three independent random seeds on the six-task
RoboTwin 2.0 subset described in Appendix~\ref{app:ablation_studies}, following
the same evaluation protocol. Table~\ref{tab:app_training_seeds} reports mean
success rates and sample standard deviations. PriorVLA outperforms FFT under
every training seed on both Easy and Hard.

\begin{table}[!t]
\centering
\small
\setlength{\tabcolsep}{7pt}
\caption{
RoboTwin 2.0 results across three training seeds. Mean success rates (\%) and
sample standard deviations are reported on the six-task subset. Best results
are shown in bold, and the PriorVLA row is shaded.
}
\label{tab:app_training_seeds}
\begin{tabular}{lcc}
\toprule
Method & Easy & Hard \\
\midrule
FFT & $64.6 \pm 1.4$ & $33.7 \pm 0.4$ \\
\rowcolor{gray!12} \textbf{PriorVLA (Ours)} & $\mathbf{77.3 \pm 0.1}$ & $\mathbf{49.0 \pm 0.6}$ \\
\bottomrule
\end{tabular}
\end{table}

\subsection{Inference Cost}
\label{app:inference_cost}

We measure model-only inference on an NVIDIA A100 using BF16, batch size 1,
and 10 denoising steps. Latency is averaged over 300 action chunks after
warm-up, with 50 actions per chunk.

\begin{table}[!t]
\centering
\small
\setlength{\tabcolsep}{7pt}
\caption{
Model-only inference cost. Each action chunk contains 50 actions.
}
\label{tab:app_inference_cost}
\begin{tabular}{lccc}
\toprule
Method & Latency (ms/chunk) & Throughput (chunks/s) & Peak VRAM (GiB) \\
\midrule
$\pi_{0.5}$ & 88.59 & 11.29 & 6.58 \\
\rowcolor{gray!12} PriorVLA & 124.51 & 8.03 & 7.46 \\
\bottomrule
\end{tabular}
\end{table}

As shown in Table~\ref{tab:app_inference_cost}, PriorVLA increases inference latency by 40.5\% and peak GPU memory by 0.88 GiB.
The frozen Prior Expert is executed alongside the Adaptation Expert at each
denoising step, while vision encoder and VLM computation are shared.

\subsection{Attention Analysis}
\label{app:attention_analysis}

We analyze attention after masking and softmax over all action chunks from
eight Easy/Hard episodes across the four-task subset described in Appendix~\ref{app:ablation_studies}. We average attention
over heads, Query tokens of the same type, and the final four layers.

\begin{table}[!t]
\centering
\small
\setlength{\tabcolsep}{7pt}
\caption{
Attention allocation of Expert Queries and AE action tokens. Values are
percentages of attention assigned to each source. The final row reports the
combined attention of AE action tokens to SQ, MQ, and AQ.
}
\label{tab:app_attention_analysis}
\begin{tabular}{llr}
\toprule
Tokens & Source & Attention (\%) \\
\midrule
\multirow{3}{*}{SQ} & Image OBS & 25.4 \\
 & Text/State OBS & 40.9 \\
 & SQ self & 33.7 \\
\midrule
\multirow{2}{*}{MQ} & Frozen PE states & 55.7 \\
 & MQ self & 44.3 \\
\midrule
\multirow{4}{*}{AQ} & Direct OBS & 38.2 \\
 & SQ & 5.6 \\
 & MQ & 4.3 \\
 & AQ self + AE action tokens & 51.8 \\
\midrule
AE action tokens & SQ + MQ + AQ & 12.8 \\
\bottomrule
\end{tabular}
\end{table}

As shown in Table~\ref{tab:app_attention_analysis}, SQ attends to both image
and text/state context, while MQ reads frozen PE states. AQ assigns most
attention to direct OBS and action states, with SQ and MQ providing an
additional input path.

\section{Task Details}
\label{app:task_details}

To ensure comprehensive evaluation across simulation and real-world settings, we summarize the tasks and language instructions used in our experiments. The evaluation covers real-world manipulation on Franka and AC-One, single-arm simulation in LIBERO, and bimanual simulation in RoboTwin 2.0.

\subsection{Real-World Tasks}
\label{app:task_details_real_world}

The real-world benchmark consists of eight manipulation tasks across a single-arm Franka robot and a dual-arm AC-One platform. These tasks cover precise insertion, object picking, stacking, sweeping, rearrangement, and deformable-object manipulation. Each task is evaluated under both ID and OOD conditions. ID trials use the nominal task setup, while OOD trials perturb lighting, background objects, initial object locations, and table height. Table~\ref{tab:app_real_world_task_details_candidate} summarizes the platform assignment and representative instruction for each real-world task.

\begin{table}[!ht]
\centering
\small
\setlength{\tabcolsep}{4pt}
\renewcommand{\arraystretch}{1.12}
\caption{
Real-world task details. We list the task prompts used for real-world evaluation.
}
\label{tab:app_real_world_task_details_candidate}
\begin{tabular}{p{0.18\linewidth}p{0.14\linewidth}p{0.58\linewidth}}
\toprule
Task & Platform & Prompt \\
\midrule
Place Ring & Franka & Pick up the ring and place it onto the vertical peg. \\
Insert Peg & Franka & Pick up the object and insert it into the matching slot. \\
Pick Object & Franka & Pick up the target object and place it into the basket. \\
Stack Blocks & Franka & Pick up one block and place it on top of the other block. \\
Stack Bowls & AC-One & Pick up the two bowls and stack one bowl on top of the other. \\
Sweep Blocks & AC-One & Use one arm to hold the container and the other arm to push all objects into it. \\
Arrange Cups & AC-One & Use the left arm to stack two paper cups, press them together, hand them to the right arm, and place them into the rack. \\
Hang Towel & AC-One & Use one arm to hold the hanger and the other arm to place the towel onto it, then return the hanger to its position. \\
\bottomrule
\end{tabular}
\end{table}

\subsection{LIBERO Tasks}
\label{app:task_details_libero}

The LIBERO benchmark evaluates single-arm manipulation in simulation with a Franka robot. We use four suites: Spatial, Object, Goal, and Long. LIBERO-Spatial consists of tasks where the same object is placed across varying target positions, stressing spatial grounding and positional generalization. LIBERO-Object focuses on handling diverse objects within similar scene layouts, testing object-centric recognition and manipulation. LIBERO-Goal contains heterogeneous operations such as opening containers, placing objects into or onto target regions, and interacting with appliances, performed in a shared environment. LIBERO-Long introduces extended tasks that require multiple sub-goals across different scenes, providing a more challenging setting for long-horizon execution and error accumulation. Together, these suites evaluate whether the policy can follow language instructions across spatial variation, object variation, goal variation, and long-horizon task composition. Table~\ref{tab:app_libero_task_details_candidate} summarizes the instruction templates and task counts for the four LIBERO suites.

\begin{table}[!ht]
\centering
\small
\setlength{\tabcolsep}{4pt}
\renewcommand{\arraystretch}{1.12}
\caption{
LIBERO task details. We list the language instruction templates and the total number of tasks per suite.
}
\label{tab:app_libero_task_details_candidate}
\begin{tabular}{p{0.11\linewidth}|p{0.72\linewidth}|p{0.08\linewidth}}
\toprule
Suite & Language Instruction Templates & \#Tasks \\
\midrule
Spatial & pick up the \textit{OBJ SPATIAL\_REL} and place it on the \textit{TARGET} & 10 \\
\midrule
Object & pick up the \textit{FOOD} and place it in the \textit{CONTAINER} & 10 \\
\midrule
Goal & open/close the \textit{CONTAINER} & 10 \\
 & open the \textit{DRAWER} and put the \textit{OBJ} inside & \\
 & put the \textit{OBJ} on/in the \textit{TARGET} & \\
 & push the \textit{OBJ} to the \textit{POSITION} of the \textit{TARGET} & \\
 & turn on the \textit{APPLIANCE} & \\
\midrule
Long-10 & put both \textit{OBJ1} and \textit{OBJ2} in the \textit{CONTAINER} & 10 \\
 & turn on the \textit{APPLIANCE} and put the \textit{OBJ} on it & \\
 & put the \textit{OBJ} in the \textit{CONTAINER/APPLIANCE} and close it & \\
 & place \textit{OBJ1} on \textit{TARGET1} and \textit{OBJ2} on \textit{TARGET2}/at \textit{REL} of \textit{TARGET2} & \\
 & pick up the \textit{OBJ} and place it in the caddy \textit{COMPARTMENT} & \\
\bottomrule
\end{tabular}
\end{table}

\subsection{RoboTwin 2.0 Tasks}
\label{app:task_details_robotwin2}

The RoboTwin 2.0 benchmark evaluates bimanual manipulation in simulation with the Aloha-AgileX robot. We evaluate 13 tasks from the official benchmark, covering grasping, handover, lifting, articulated-object interaction, object placement, container manipulation, and bimanual stacking. RoboTwin 2.0 also provides Seen and UnSeen language settings. Seen instructions are sampled from the training prompt pool, while UnSeen instructions use held-out paraphrases that describe the same task with different wording. Table~\ref{tab:app_robotwin2_prompt_example} shows one Open Laptop instruction from each split as a representative example. In addition to language variation, RoboTwin 2.0 is evaluated under Easy and Hard scene settings, where Easy corresponds to the clean ID setting and Hard corresponds to the randomized OOD setting.

On the six-task subset described in Appendix~\ref{app:ablation_studies}, PriorVLA achieves 78.5\% Easy and 48.5\% Hard success with Seen instructions, compared with 77.3\% and 48.3\% with UnSeen instructions. FFT and LoRA show similarly small differences.

\begin{table}[!ht]
\centering
\small
\setlength{\tabcolsep}{5pt}
\renewcommand{\arraystretch}{1.12}
\caption{
RoboTwin 2.0 Open Laptop language-prompt example. We show one Seen instruction and one UnSeen instruction to illustrate the language split used in evaluation.
}
\label{tab:app_robotwin2_prompt_example}
\begin{tabular}{p{0.16\linewidth}p{0.64\linewidth}}
\toprule
Language split & Example instruction \\
\midrule
Seen & Lift the laptop with silver base open with the right arm. \\
\midrule
UnSeen & Open the hinged screen laptop by lifting its lid. \\
\bottomrule
\end{tabular}
\end{table}

The full RoboTwin 2.0 task set used in our experiments contains the following 13 task names: \textit{Stack Bowls Two}, \textit{Stack Blocks Two}, \textit{Put Object Cabinet}, \textit{Put Bottles Dustbin}, \textit{Place Phone Stand}, \textit{Place Dual Shoes}, \textit{Place Object Basket}, \textit{Pick Dual Bottles}, \textit{Open Laptop}, \textit{Move Can Pot}, \textit{Lift Pot}, \textit{Handover Mic}, and \textit{Grab Roller}.

\section{Qualitative Results}
\label{app:qualitative_results}

We provide qualitative rollout examples to complement the quantitative results
in the main paper. These visualizations illustrate representative policy
executions in both real-world and simulation environments, and help show how the
policy behaves under in-distribution (ID) and out-of-distribution (OOD)
conditions. We first summarize what each qualitative figure contains, and then
present the figures as a visual gallery.

\begin{figure*}[p]
    \centering
    \includegraphics[
        width=\textwidth,
        trim={105pt 300pt 100pt 10pt},
        clip
    ]{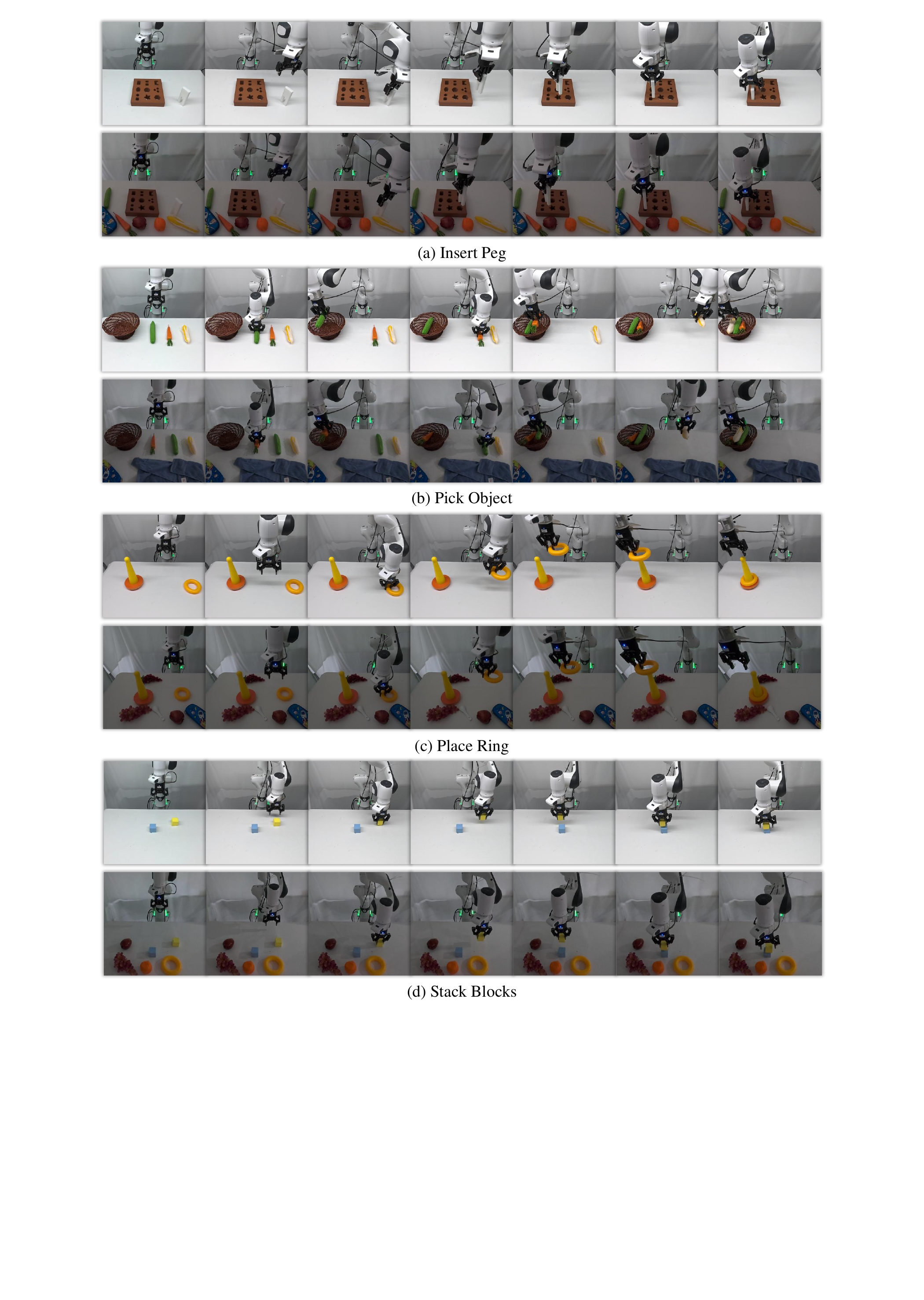}
    \caption{
    \textbf{Qualitative real-world results on the Franka platform.}
    We show rollout snapshots for four representative tasks under both
    in-distribution (ID) and out-of-distribution (OOD) evaluation. These examples
    illustrate representative task progression under nominal and perturbed
    real-world conditions.
    }
    \label{fig:qual_real_franka}
\end{figure*}

\begin{figure*}[p]
    \centering
    \includegraphics[
        width=\textwidth,
        trim={105pt 300pt 100pt 10pt},
        clip
    ]{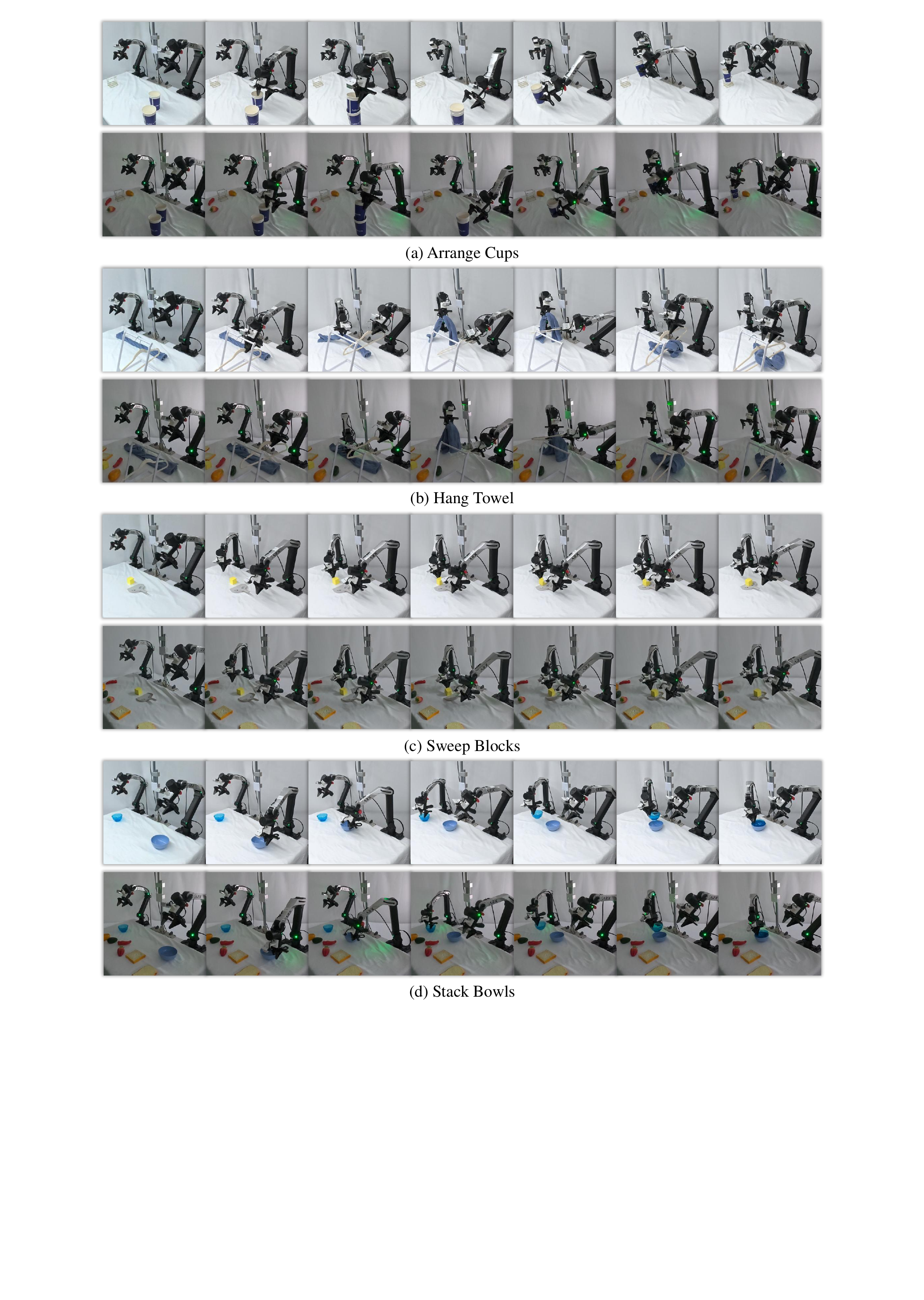}
    \caption{
    \textbf{Qualitative real-world results on the AC-One platform.}
    We show rollout snapshots for four representative dual-arm tasks under both
    in-distribution (ID) and out-of-distribution (OOD) evaluation. These examples
    illustrate representative task progression on the dual-arm platform.
    }
    \label{fig:qual_real_acone}
\end{figure*}

\begin{figure*}[p]
    \centering
    \includegraphics[
        width=\textwidth,
        trim={105pt 470pt 100pt 10pt},
        clip
    ]{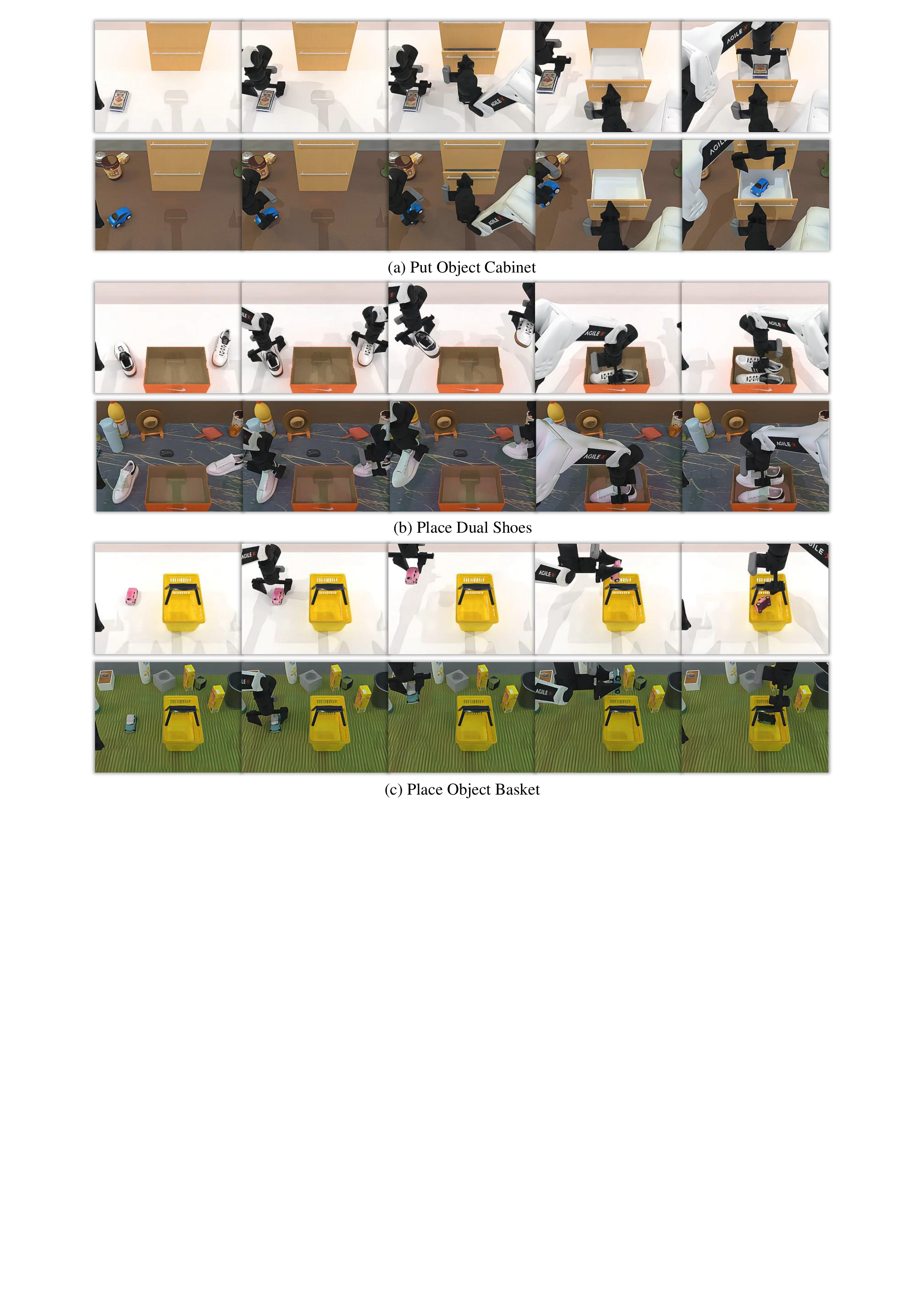}
    \caption{
    \textbf{Qualitative simulation results on RoboTwin 2.0 (Part I).}
    We show additional representative rollout snapshots under ID/Easy and
    OOD/Hard evaluation.
    }
    \label{fig:qual_sim_robotwin2_part1}
\end{figure*}

\begin{figure*}[p]
    \centering
    \includegraphics[
        width=\textwidth,
        trim={105pt 470pt 100pt 10pt},
        clip
    ]{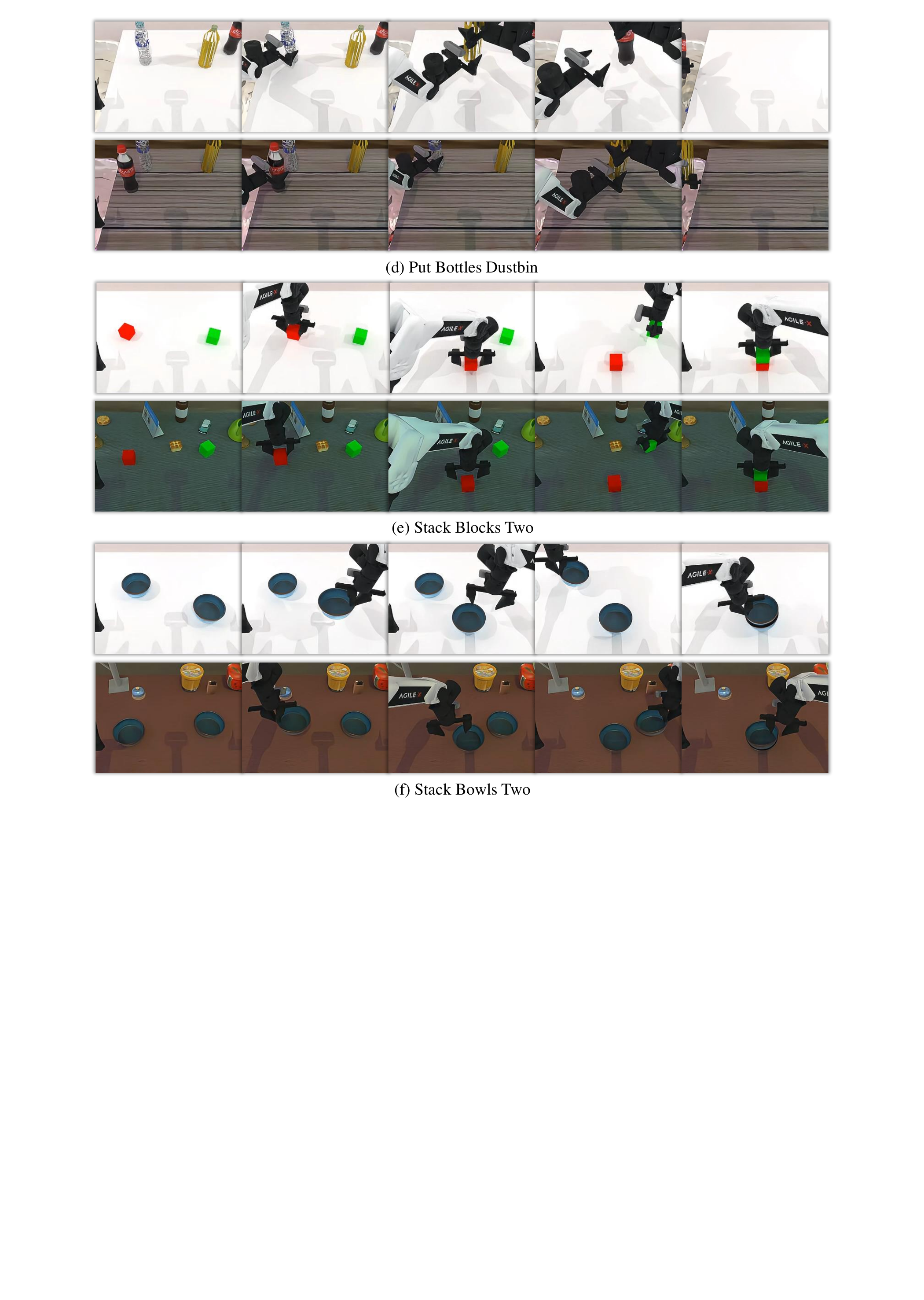}
    \caption{
    \textbf{Qualitative simulation results on RoboTwin 2.0 (Part II).}
    We show additional representative rollout snapshots under ID/Easy and
    OOD/Hard evaluation.
    }
    \label{fig:qual_sim_robotwin2_part2}
\end{figure*}

\begin{figure*}[p]
    \centering
    \includegraphics[
        width=\textwidth,
        trim={110pt 800pt 110pt 10pt},
        clip
    ]{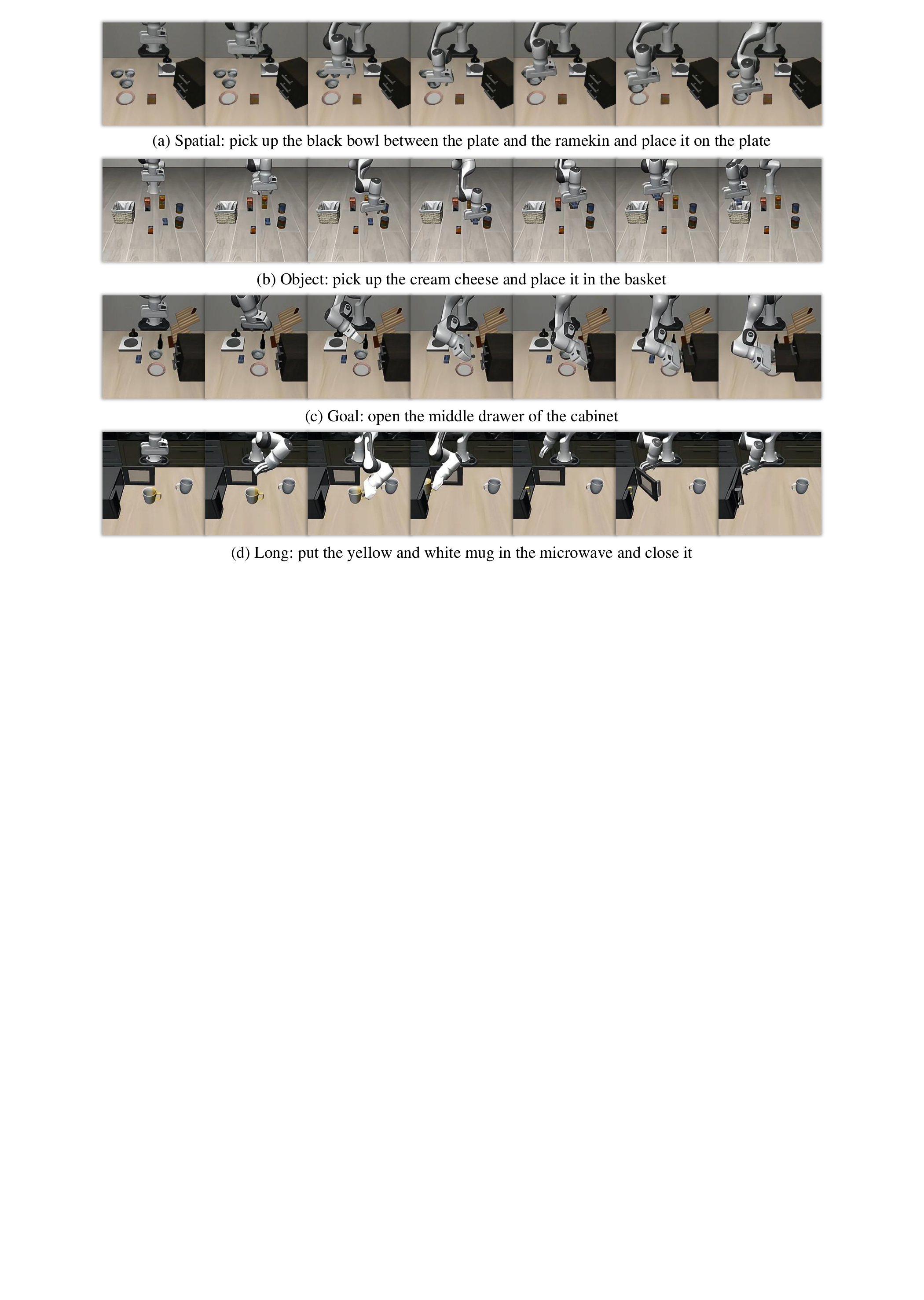}
    \caption{
    \textbf{Qualitative simulation results on LIBERO.}
    We show one representative task from each of the four suites. Since LIBERO
    is evaluated under the standard benchmark setting, only in-distribution
    rollouts are shown.
    }
    \label{fig:qual_sim_libero}
\end{figure*}

The qualitative rollout figures are organized as follows:
(1) Fig.~\ref{fig:qual_real_franka} shows Franka real-world rollouts across four
tasks, with both ID and OOD trials.
(2) Fig.~\ref{fig:qual_real_acone} shows AC-One real-world rollouts across four
dual-arm tasks, with both ID and OOD trials.
(3) Fig.~\ref{fig:qual_sim_robotwin2_part1} shows the first set of RoboTwin 2.0
simulation rollouts under Easy/ID and Hard/OOD evaluation.
(4) Fig.~\ref{fig:qual_sim_robotwin2_part2} shows additional RoboTwin 2.0
simulation rollouts under Easy/ID and Hard/OOD evaluation.
(5) Fig.~\ref{fig:qual_sim_libero} shows representative LIBERO rollouts, one
from each evaluated suite.


\clearpage

\section{VQA Analysis}
\label{app:vqa_analysis}

To further examine whether downstream adaptation preserves visual-language reasoning ability, we conduct a small set of qualitative VQA probes. These probes are not used as an additional benchmark or training objective; instead, they provide diagnostic evidence about whether a model retains basic vision-language understanding after robot-task adaptation. The prompts cover both general visual understanding, such as object recognition, digit recognition, counting, and scene description, and robot-relevant questions, such as predicting the next subtask from the current manipulation scene.

For a fair comparison, we disable Scene Queries during this VQA probing stage. Scene Queries are a PriorVLA-specific interface designed to capture task-relevant scene priors for action generation. Keeping them active during VQA could introduce an additional query pathway that is not present in the pretrained model or the full fine-tuning baseline. We therefore evaluate all models through the shared vision-language pathway, so that the comparison reflects how much visual-language ability is preserved after adaptation rather than whether an additional query interface can assist VQA.

\begin{figure}[!t]
    \centering
    \setlength{\abovecaptionskip}{2pt}
    \setlength{\belowcaptionskip}{0pt}
    \includegraphics[
        width=\linewidth,
        trim={250pt 85pt 250pt 130pt},
        clip
    ]{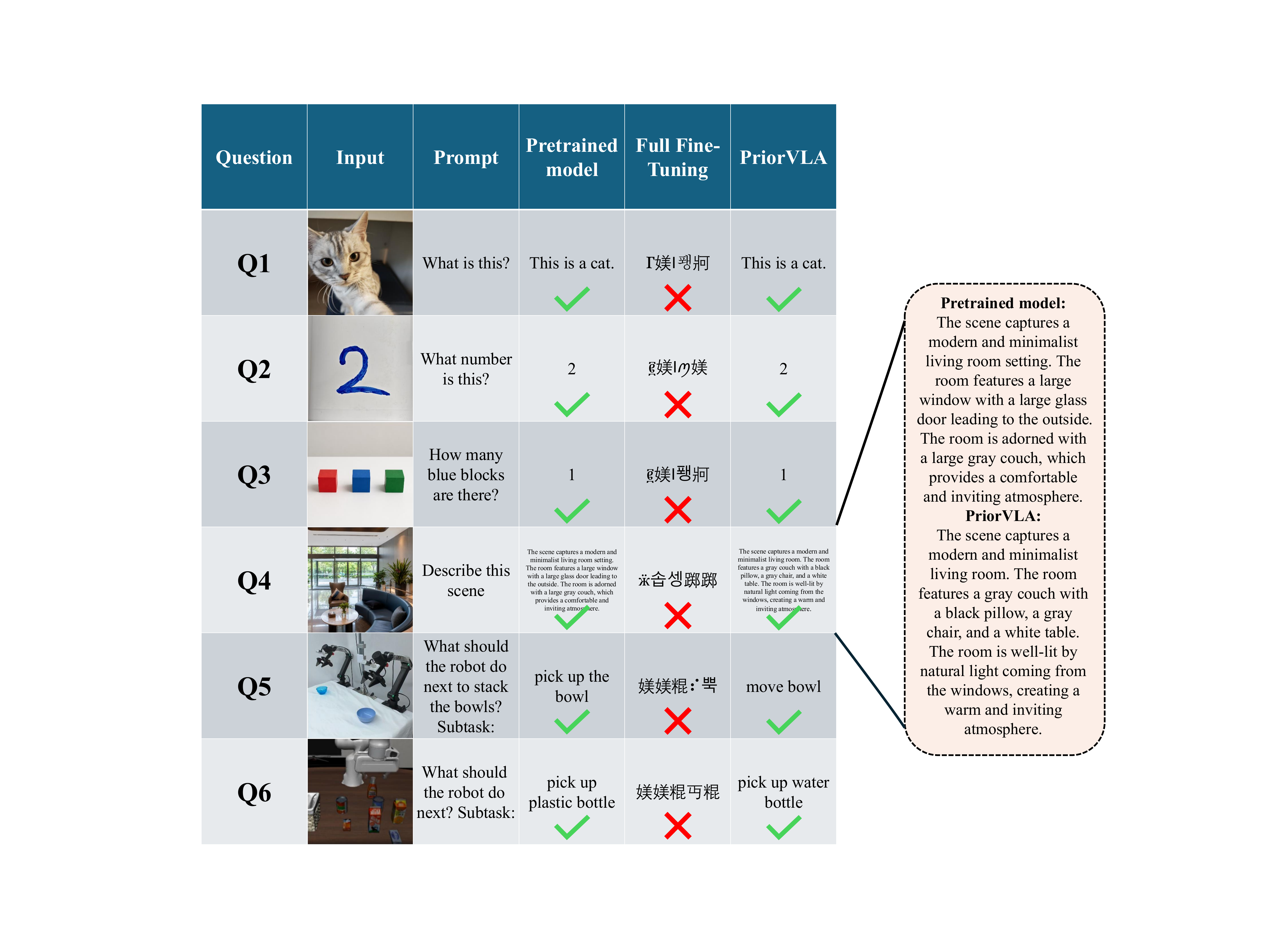}
    \vspace{-8mm}
    \caption{
    Qualitative VQA examples after adaptation.
    The probes include general visual recognition, counting, scene description, and robot-relevant subtask questions.
    During VQA probing, Scene Queries are disabled to ensure that PriorVLA does not use an additional VQA-specific query pathway.
    Full fine-tuning produces non-semantic outputs in these examples, while PriorVLA preserves coherent visual-language responses similar to the pretrained model.
    }
    \label{fig:app_vqa_examples}
    \vspace{-2mm}
\end{figure}

Figure~\ref{fig:app_vqa_examples} shows representative VQA results. The pretrained model produces coherent answers for all probes, confirming that the base VLA retains general visual-language understanding before downstream robot-task adaptation. Full fine-tuning, however, produces non-semantic and unreadable responses in these examples. This suggests that directly updating the full model on action-supervision data can disturb the language-generation behavior of the pretrained VLM, even when the downstream objective is not designed to modify VQA ability. In contrast, PriorVLA preserves coherent VQA responses after adaptation. It correctly recognizes common objects and digits, counts scene elements, generates a meaningful scene description, and produces reasonable robot-relevant subtask answers.

These results are consistent with the motivation of PriorVLA. Full fine-tuning uses the pretrained model primarily as an initialization for downstream action prediction, which can shift pretrained visual-language representations toward the narrow robot demonstration distribution. PriorVLA instead separates prior preservation from downstream specialization. The frozen Prior Expert preserves action-side motor priors, the VLM core is kept largely frozen while the vision encoder and adaptation
branch specialize to downstream actions. As a result, PriorVLA can improve downstream robot performance while retaining more of the pretrained model's vision-language functionality. Although the VQA probes are qualitative and not intended as a standalone benchmark, they provide supporting evidence that PriorVLA better preserves general-purpose visual-language priors during robot-task adaptation.

\section{Additional Discussion}
\label{app:additional_discussion}

\subsection{Why Prior Preservation Helps OOD Generalization}
\label{app:why_ood}

Out-of-distribution generalization in downstream VLA adaptation is challenging
because downstream demonstrations often cover only a limited range of visual
conditions, object configurations, and workspace layouts. Full fine-tuning
updates the pretrained action-generation pathway directly with this restricted
supervision. While this can improve performance on the training distribution, it
may also shift pretrained representations and action priors toward narrow
downstream patterns, reducing robustness when evaluation scenes contain unseen
lighting, backgrounds, object poses, clutter, or workspace changes.

PriorVLA mitigates this issue by separating prior preservation from downstream
specialization. The frozen Prior Expert preserves the pretrained
action-generation pathway and serves as a read-only source of motor-prior
representations. The Adaptation Expert, initialized from the same pretrained
action expert, specializes to downstream action generation. Expert Queries then
provide learnable interfaces for using preserved priors: Scene Queries capture
task-relevant scene priors from the VLM, Motor Queries capture motor-prior
representations from the frozen Prior Expert, and Action Queries integrate these
priors inside the trainable Adaptation Expert.

This design is especially useful under distribution shift. When visual scenes or
object configurations differ from the downstream training data, the policy cannot
rely only on patterns observed in task-specific demonstrations. It can instead
reuse priors learned during large-scale pretraining, such as object-centric
visual grounding, affordance-related scene understanding, and general motor
regularities. The RoboTwin 2.0-Hard and real-world OOD results are consistent with
this interpretation: PriorVLA shows larger gains under OOD evaluation than under
easier ID settings.

\subsection{Why Prior Preservation Helps Few-Shot Adaptation}
\label{app:why_few_shot}

Few-shot adaptation further amplifies the risk of over-specialization because a
small number of downstream demonstrations covers only a narrow subset of initial
states, visual appearances, object poses, and successful action trajectories. In
this regime, full fine-tuning can make the adapted policy sensitive to incidental
correlations in the few-shot data, rather than learning task-relevant structure
that transfers across evaluation conditions.

PriorVLA reduces this risk by separating a frozen prior source from a trainable
adaptation branch. The frozen Prior Expert maintains a stable source of
motor-prior features, while the Adaptation Expert learns the downstream action
mapping. Expert Queries serve as learnable interfaces for selecting and
integrating preserved priors, rather than replacing the pretrained prior itself.
As a result, the model does not need to recover all relevant visual and motor
structure from few-shot data alone; it can reuse pretrained priors and learn how
to integrate them for the downstream task.

The few-shot results are consistent with this view. PriorVLA improves over full
fine-tuning in both simulation and real-world few-shot settings, with especially
strong gains under OOD evaluation. This suggests that the advantage of PriorVLA
comes not only from parameter efficiency, but also from preserving and reusing
pretrained knowledge when downstream supervision is limited.

\subsection{Broader Impacts}
\label{app:broader_impacts}

PriorVLA aims to improve data-efficient adaptation of pretrained robot policies.
Its potential positive impact is to reduce the amount of task-specific robot data
needed for downstream adaptation, which may lower the cost of developing robot
systems and improve robustness under moderate distribution shifts.

Potential negative impacts may arise if stronger robot policies are deployed
without sufficient validation or supervision. Failures under unseen scenes,
contact-rich interactions, or unsafe environments could lead to physical damage
or unintended robot behavior. The method could also be misused in inappropriate
or unauthorized automation settings if deployed without task and environment
restrictions. In our work, real-world experiments are conducted in controlled lab
settings with human supervision, task-specific evaluation, and standard safety
procedures. We do not claim that the policy is ready for unsupervised deployment.

\subsection{Real-World Deployment Notes}
\label{app:real_robot_deployment_notes}

Real-robot experiments are conducted under human supervision and with standard safety procedures. A trial is counted as a failure if the robot requires human intervention, triggers a safety stop, violates the task constraints, or fails to satisfy the success criterion before timeout. These criteria are applied consistently across methods.

During deployment, the policy predicts an action chunk of length 50 and executes the first 15 actions before replanning with a new observation. This closed-loop chunking strategy limits open-loop drift while keeping the inference frequency manageable for real-robot control. The same action-horizon and execution-horizon settings are used across the evaluated real-robot tasks.

Before each evaluation batch, the workspace is reset according to the task protocol. We check workspace boundaries, robot speed limits, emergency-stop access, camera visibility, object placement, and table clearance. For OOD evaluation, perturbations such as lighting changes, background-object changes, initial object-pose variation, and table-height changes are applied while maintaining safe operating conditions. These procedures ensure that failures are attributed to policy behavior under the specified evaluation condition rather than to uncontrolled hardware or reset inconsistencies.

\subsection{Existing Assets and Licenses}
\label{app:existing_assets_licenses}

We build on and evaluate with existing open-source assets, and follow their
corresponding licenses and terms of use. Table~\ref{tab:app_existing_assets}
summarizes the main external assets used in this work.

\clearpage  

\begin{table}[p]  
\centering
\small
\setlength{\tabcolsep}{4pt}
\renewcommand{\arraystretch}{1.12}
\caption{
\textbf{Existing assets used in PriorVLA.} We cite the corresponding papers and
respect the licenses and terms of use of each asset.
}
\label{tab:app_existing_assets}
\begin{tabular}{
@{}
>{\raggedright\arraybackslash}p{0.18\linewidth}
>{\raggedright\arraybackslash}p{0.42\linewidth}
>{\raggedright\arraybackslash}p{0.32\linewidth}
@{}}
\toprule
\textbf{Asset} & \textbf{Use in this work} & \textbf{License / Terms} \\
\midrule

OpenPI ($\pi_{0.5}$)
& Base VLA implementation and pretrained backbone used for adaptation.
& Apache-2.0 for the OpenPI codebase; model/checkpoint use follows the upstream terms, including the Gemma Terms of Use where applicable. \\
\midrule

RoboTwin 2.0
& Simulation benchmark, official demonstrations, and data-generation pipeline for few-shot and large-data regimes.
& MIT License. \\
\midrule

LIBERO
& Simulation benchmark and public task demonstrations for the four LIBERO suites.
& MIT License for code; CC BY 4.0 for datasets. \\
\midrule

LeRobot
& LeRobot data format and tooling used for converting and organizing demonstrations.
& Apache-2.0 License. \\

\bottomrule
\end{tabular}
\end{table}

\clearpage

\end{document}